\documentclass[sigconf]{acmart}
\usepackage{multirow}
\usepackage{makecell}
\usepackage{multirow}
\usepackage{mathrsfs}
\usepackage[ruled]{algorithm2e}
\usepackage{cellspace}
\newtheorem{theorem}{Theorem}  
  
\newtheorem{lemma}{Lemma}

\usepackage{makecell}
\usepackage{bm}
\usepackage{graphicx}
\usepackage{paralist}
\def\*#1{\bm{#1}}
\def\&#1{\mathcal{#1}}
\def\E{\mathbb{E}}
\def\!#1{\vec{\bm{#1}}}
\def\BibTeX{{\rm B\kern-.05em{\sc i\kern-.025em b}\kern-.08emT\kern-.1667em\lower.7ex\hbox{E}\kern-.125emX}}
\definecolor{comment}{RGB}{70, 150, 60}

\copyrightyear{2020}
\acmYear{2020}
\setcopyright{acmcopyright}
\acmConference[KDD '20]{Proceedings of the 26th ACM SIGKDD Conference on Knowledge Discovery and Data Mining}{August 23--27, 2020}{Virtual Event, CA, USA}
\acmBooktitle{Proceedings of the 26th ACM SIGKDD Conference on Knowledge Discovery and Data Mining (KDD '20), August 23--27, 2020, Virtual Event, CA, USA}
\acmPrice{15.00}
\acmDOI{10.1145/3394486.3403218}
\acmISBN{978-1-4503-7998-4/20/08}

\begin{document}
\fancyhead{}

\title{Understanding Negative Sampling in Graph Representation Learning}

\author[Z. Yang*, M. Ding*, C. Zhou, H. Yang, J. Zhou, J. Tang]{Zhen Yang$^{*\dagger}$, Ming Ding$^{*\dagger}$, Chang Zhou$^{\ddagger}$, Hongxia Yang$^{\ddagger}$, Jingren Zhou$^{\ddagger}$, Jie Tang$^{\dagger \S}$}
\affiliation{
    $^\dagger$ Department of Computer Science and Technology, Tsinghua University
}
\affiliation{
    $^\ddagger$ DAMO Academy, Alibaba Group
}
\email{zheny2751@gmail.com, dm18@mails.tsinghua.edu.cn}
\email{{ericzhou.zc,yang.yhx,jingren.zhou}@alibaba-inc.com}
\email{jietang@tsinghua.edu.cn}

\renewcommand{\authors}{Zhen Yang, Ming Ding, Chang Zhou, Hongxia Yang, Jingren Zhou, Jie Tang}






\newcommand{\model}{MCNS}
\newcommand{\smodel}{\model\space}
\begin{abstract}
\renewcommand{\thefootnote}{\fnsymbol{footnote}}
\footnotetext[1]{Equal contribution. Ming Ding proposed the theories and method. Zhen Yang conducted the experiments. Codes are available at \url{https://github.com/zyang-16/MCNS}.}
\footnotetext[4]{Corresponding Authors.}
\renewcommand{\thefootnote}{\arabic{footnote}}

Graph representation learning has been extensively studied in recent years, in which sampling is a critical point. Prior arts usually focus on sampling positive node pairs, while the strategy for negative sampling is left insufficiently explored. To bridge the gap, we systematically analyze the role of negative sampling from the perspectives of both objective and risk, theoretically demonstrating that negative sampling is as important as positive sampling in determining the optimization objective and the resulted variance. To the best of our knowledge, we are the first to derive the theory and quantify that a nice negative sampling distribution is $p_n(u|v)\propto p_d(u|v)^\alpha, 0 < \alpha < 1$. With the guidance of the theory, we propose \model, approximating the positive distribution with self-contrast approximation and accelerating negative sampling by Metropolis-Hastings. We evaluate our method on 5 datasets that cover extensive downstream graph learning tasks, including link prediction, node classification and recommendation, on a total of 19 experimental settings. These relatively comprehensive experimental results demonstrate its robustness and superiorities.

\end{abstract}

%


%


%
\maketitle

\section{Introduction} \label{sec:intro}

\begin{figure*}[htbp]
  \centering
  \includegraphics[width=\textwidth]{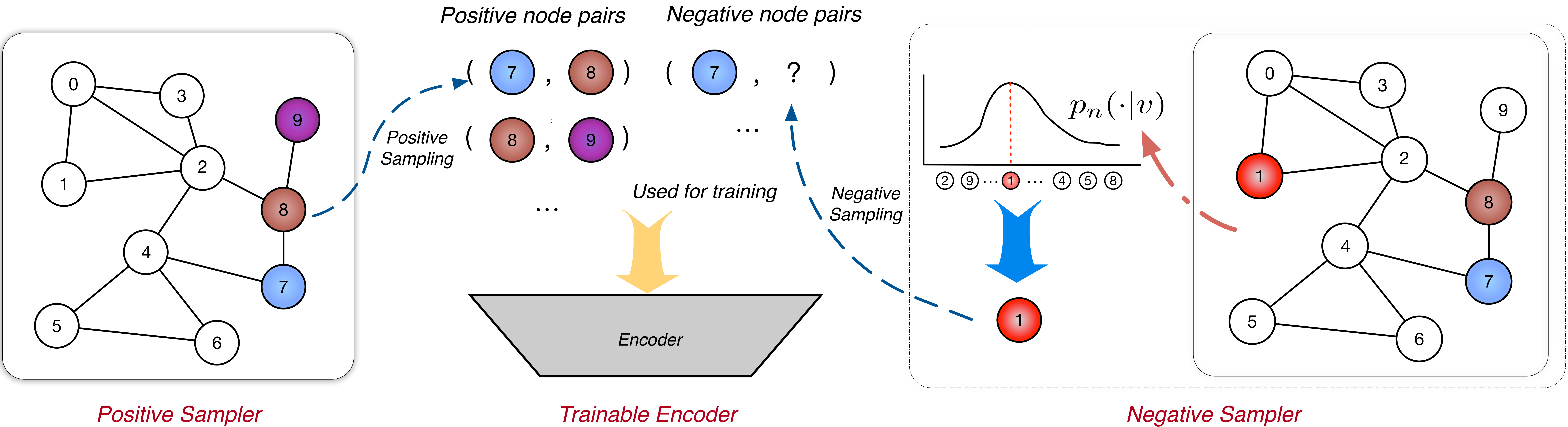}
  \vspace*{-0.4cm}
  \caption{The SampledNCE framework. Positive pairs are sampled implicitly or explicitly according to the graph representation methods, while negative pairs are from a pre-defined distribution, both composing the training data of contrastive learning.}
  \label{example}
  \vspace*{-0.3cm}
\end{figure*}

Recent years have seen the \emph{graph representation learning} gradually stepping into the spotlight of data mining research. 
The mainstream graph representation learning algorithms include traditional Network Embedding methods (e.g. DeepWalk~\cite{deepwalk}, LINE~\cite{line}) and Graph Neural Networks (e.g. GCN~\cite{gcn}, GraphSAGE~\cite{graphsage}), although the latter sometimes are trained end-to-end in classification tasks.
These methods transform nodes in the graph into low-dimensional vectors to facilitate a myriad of downstream tasks, such as node classification~\cite{deepwalk}, link prediction~\cite{zhou2017scalable} and recommendation~\cite{pinsage}.

Most graph representation learning can be unified within a \emph{Sampled Noise Contrastive Estimation} (SampledNCE) framework (\S~\ref{sec:framework}), comprising of a trainable encoder to generate node embeddings, a positive sampler and a negative sampler to sample positive and negative nodes respectively for any given node. 
The encoder is trained to distinguish positive and negative pairs based on the inner products of their embeddings (See Figure~\ref{example} for illustration). 

Massive network embedding works have investigated good criteria to sample positive node pairs, such as by random walk~\cite{deepwalk}, the second-order proximity~\cite{line}, community structure~\cite{wang2017community}, etc. 
Meanwhile, the line of GNN papers focuses on advanced encoders, evolving from basic graph convolution layers~\cite{gcn} to attention-based~\cite{velivckovic2017graph}, GAN-based~\cite{ding2018semi}, sampled aggregator~\cite{graphsage} and WL kernel-based~\cite{xu2018powerful} layers.
However, the strategy for negative sampling is relatively unexplored.

Negative sampling~\cite{word2vec} is firstly proposed to serve as a simplified version of noise contrastive estimation~\cite{mnih2013learning, gutmann2012noise}, which is an efficient way to compute the partition function of an unnormalized distribution to accelerate the training of word2vec. \citet{word2vec} set the negative sampling distribution proportional to the $3/4$ power of degree by tuning the power parameter. Most later works on network embedding~\cite{deepwalk, line} readily keep this setting. However, results in several papers~\cite{zhang2018gneg,caselles2018word2vec, pinsage} demonstrate that not only the distribution of negative sampling has a huge influence on the performance, but also the best choice varies largely on different datasets~\cite{caselles2018word2vec}.

To the best of our knowledge, few papers systematically analyzed or discussed negative sampling in graph representation learning. In this paper, we first examine the role of negative sampling from the perspectives of objective and risk. Negative sampling is as important as positive sampling to determine the optimization objective and reduce the variance of the estimation in real-world graph data. 
According to our theory, the negative sampling distribution should be \textbf{positively but sub-linearly correlated} to their positive sampling distribution. Although our theory contradicts with the intuition ``positively sampling nearby nodes and negatively sampling the faraway nodes'', it accounts for the observation in previous works~\cite{pinsage, zhang2018gneg, gao2018self-paced} where additional \emph{hard} negative samples improve performance. 

We propose an effective and scalable negative sampling strategy, \textbf{M}arkov \textbf{c}hain Monte Carlo \textbf{N}egative \textbf{S}ampling (\model), which applies our theory with an approximated positive distribution based on current embeddings. To reduce the time complexity, we leverage a special Metropolis-Hastings algorithm~\cite{metropolis1953equation} for sampling. Most graphs naturally satisfy the assumption that adjacent nodes share similar positive distributions, thus the Markov chain from neighbors can be continued to skip the \emph{burn-in} period and finally accelerates the sampling process without degradations in performance. 

We evaluate \smodel on three most general graph-related tasks, link prediction, node classification and recommendation. Regardless of the network embedding or GNN model used, significant improvements are shown by replacing the original negative sampler to \model.
We also collect many negative sampling strategies from information retrieval~\cite{weston2011wsabie,wang2017irgan}, ranking in recommendation~\cite{hu2008collaborative, dns} and knowledge graph embeddings~\cite{cai2017kbgan, zhang2019nscaching}. 
Our experiments in five datasets demonstrate that \smodel stably outperforms these negative sampling strategies.
\section{Framework}
\label{sec:framework}

\begin{algorithm}[htb]
\caption{The Sampled Noise Contrastive Estimation Framework}\label{framework}
\KwIn{
Graph $G=(V, E)$, batch size $m$, Encoder ${E}_\theta$, \\ Positive Sampling Distribution $\hat{p_d}$, \\Negative Sampling Distribution $p_n$, \\ Number of Negative Samples $k$.
} 
\Repeat{Convergence}{ 
    Sample $m$ nodes from $p(v) \propto deg(v)$ \\ 
    Sample $1$ positive node from $\hat{p_d}(u|v)$ for each node $v$\\
    Sample $k$ negative nodes from $p_n(u'|v)$ for each node $v$\\
    Initialize loss $\&L$ as zero\\
    \For{each node $v$}{
        \For{each positive node $u$ for $v$}{
        $\&L = \&L -\log \sigma(E_\theta(u)\cdot E_\theta(v))$
        }
        \For{each negative node $u'$ for $v$}{
         $\&L = \&L -\log (1-\sigma(E_\theta(u')\cdot E_\theta(v)))$
        }
    }
    Update $\theta$ by descending the gradients $\nabla_\theta\&L$
}
\end{algorithm}

In this section, we introduce the preliminaries to understand negative sampling, including the unique embedding transformation, the SampledNCE framework and how the traditional network embeddings and GNNs are unified into the framework.

\subsection{Unique Embedding Transformation}
Many network embedding algorithms, such as DeepWalk~\cite{deepwalk}, LINE~\cite{line} and node2vec~\cite{node2vec}, actually assign each node two embeddings~\footnote{Others, for example GraphSAGE~\cite{graphsage} use a unique embedding.}, node embedding and contextual embedding. When sampled as context, the contextual embedding, instead of its node embedding is used for inner product. This implementation technique stems from word2vec~\cite{word2vec}, though sometimes omitted by following papers, and improves performance by utilizing the asymmetry of the contextual nodes and central nodes.

We can split each node $i$ into central node $v_i$ and contextual node $u_i$, and each edge $i\rightarrow j$ becomes $v_i\rightarrow u_j$. Running those network embedding algorithms on the transformed bipartite graph is strictly equivalent to running on the original graph.\footnote{For bipartite graphs, always take nodes from $V$ part as central node and sample negative nodes from $U$ part.} According to the above analysis, we only need to anaylze graphs with unique embeddings without loss of generality.

\subsection{The SampledNCE Framework}\label{unify}
We unify a large part of graph representation learning  algorithms into the general SampledNCE framework, shown in Algorithm~\ref{framework}. The framework depends on an \textit{encoder} $E_\theta$, which can be either an embedding lookup table or a variant of GNN, to generate node embeddings. 

In the training process, $m$ positive nodes and $k\cdot m$ negative nodes are sampled. In Algorithm~\ref{framework}, we use the cross-entropy loss to optimize the inner product where $\sigma(x) = \frac{1}{1 + e^{-x}}$. And the other choices of loss function work in similar ways. In link prediction or recommendation, we recall the top $K$ nodes with largest $E_\theta(v) \cdot E_\theta(u)$ for each node $u$. In classification, we evaluate the performance of logistic regression with the node embeddings as features.

\subsection{Network Embedding and GNNs}
Edges in natural graph data can be assumed as sampled from an underlying positive distribution $p_d(u,v)$. 
However, only a few edges are observable so that numerous techniques are developed to make reasonable assumptions to better estimate $p_d(u,v)$, resulting in various $\hat{p_d}$ in different algorithms to approximate $p_d$. 

Network embedding algorithms employ various positive distributions implicitly or explicitly. LINE~\cite{line} samples positive nodes directly from adjacent nodes. DeepWalk~\cite{deepwalk} samples via random walks and implicitly defines $\hat{p_d}(u|v)$ as the stationary distribution of random walks~\cite{qiu2018network}. node2vec~\cite{node2vec} argues that homophily and structural equivalence of nodes $u$ and $v$ indicates a larger $p_d(u,v)$. These assumptions are mainly based on local smoothness and augment observed positive node pairs for training.

Graph neural networks are equipped with different encoder layers and perform implicit regularizations on positive distribution. Previous work~\cite{li2018deeper} reveals that GCN is a variant of graph Laplacian smoothing, which directly constrains the difference of the embedding of adjacent nodes. GCN, or its sampling version GraphSAGE~\cite{graphsage}, does not explicitly augment positive pairs but regularizes the output embedding by local smoothness, which can be seen as an implicit regularization of $p_d$.

\section{Understanding Negative Sampling}\label{sec:theory}
In contrast to elaborate estimation of the positive sampling distribution, negative sampling has long been overlooked. In this section, we revisit negative sampling from the perspective of objective and risk, and conclude that ``the negative sampling distribution should be positively but sub-linearly correlated to their positive sampling distribution''. 

\subsection{How does negative sampling influence the objective of embedding learning?}\label{bias}
Previous works~\cite{qiu2018network} interpret network embedding as implicit matrix factorization. We discuss a more general form where the factorized matrix is determined by estimated data distribution $\hat{p_d}(u|v)$ and negative distribution $p_n(u'|v)$. According to ~\cite{levy2014neural}~\citep{qiu2018network}, the objective function for embedding is
\begin{align} 
    J &= \E_{(u,v)\sim p_d} \log \sigma(\!{u}^\top \!{v}) + \E_{v \sim p_d(v)}[k\E_{u'\sim p_n(u'|v)} \log \sigma(-\!{u'}^{\top} \!{v})]\nonumber\\
    &= \E_{v \sim p_d(v)}[\E_{u\sim p_d(u|v)}\log \sigma(\!{u}^\top \!{v}) + k\E_{u'\sim p_n(u'|v)}\log \sigma(-\!{u'}^{\top} \!{v})],
\end{align}
\noindent where $\!u,\!v$ are embeddings for node $u$ and $v$ and $\sigma(\cdot)$ is the sigmoid function. We can separately consider each node $v$ and show the objective for optimization as follows:
\begin{theorem}\textbf{(Optimal Embedding)} \label{t1}
The optimal embedding satisfies that for each node pair $(u,v)$,  
\begin{align}
    \!u^\top \!v = -\log \frac{k\cdot p_n(u | v)}{p_d(u|v)}. \label{obj}
\end{align}
\end{theorem}
\textbf{Proof\ }
Let us consider the conditional data and the negative distribution of node $v$. To maximize $J$ is equivalent to minimize the following objective function for each $v$,
\begin{align*}
    J^{(v)} &= \E_{u\sim p_d(u|v)}\log \sigma(\!{u}^\top \!{v}) + k\E_{u'\sim p_n(u'|v)}\log \sigma(-\!{u'}^{\top} \!{v})\\
    &= \sum\limits_u  p_d(u|v) \log\sigma(\!{u}^\top \!{v}) + k\sum\limits_{u'} p_n(u'|v) \log\sigma(-\!{u'}^{\top} \!{v})\\
    &= \sum\limits_u [p_d(u|v) \log\sigma(\!{u}^\top \!{v}) + kp_n(u|v) \log\sigma(-\!{u}^\top \!{v})]\\
    &= \sum\limits_u [p_d(u|v) \log\sigma(\!{u}^\top \!{v}) + kp_n(u|v) \log\big(1-\sigma(\!{u}^\top \!{v})\big)].
\end{align*}
For each $(u,v)$ pair, if we define two Bernoulli distributions $P_{u,v}(x)$ where $P_{u,v}(x=1) = \frac{p_d(u|v)}{p_d(u|v) + k p_n(u|v)}$ and $Q_{u,v}(x)$ where $Q_{u,v}(x=1)=\sigma(\!{u}^\top \!{v})$, the equation above is simplified as follows:
\begin{align*}
    J^{(v)} &= -\sum\limits_u (p_d(u|v) + k p_n(u|v))H(P_{u,v}, Q_{u,v}),
\end{align*}
where $H(p, q)=-\sum_{x\in \mathcal{X}} p(x)\log q(x)$ is the cross entropy between two distributions. According to \textit{Gibbs Inequality}, the maximizer of $L^{(v)}$ satisfies $P = Q$ for each $(u,v)$ pair, which gives
\begin{align}
    \frac{1}{1 + e^{-\!{u}^{T} \!{v}}}  &= \frac{p_d(u|v)}{p_d(u|v) + k p_n(u|v)}\label{mterm}\\
    \!u^\top \!v &= -\log \frac{k\cdot p_n(u | v)}{p_d(u|v)}.&\tag*{\qed}\nonumber
\end{align}
The above theorem clearly shows that the positive and negative distributions $p_d$ and $p_n$ have the same level of influence on optimization. In contrast to abundant research on finding a better $\hat{p_{d}}$ to approximate $p_d$, negative distribution $p_n$ is apparently under-explored. 
\subsection{How does negative sampling influence the expected loss (risk) of embedding learning?}\label{variance}
The analysis above shows that with enough (possibly infinite) samples of edges from $p_d$, the inner product of embedding $\!u^\top\!v$ has an optimal value. In real-world data, we only have limited samples from $p_d$, which leads to a nonnegligible \textit{expected loss} (\textit{risk}). Then the loss function to minimize \textit{empirical risk}  for node $v$ becomes:
\begin{align}
    J^{(v)}_T = \frac{1}{T}\sum_{i=1}^T \log \sigma(\!{u}_i^\top \!{v}) + \frac{1}{T}\sum_{i=1}^{kT}\log \sigma(-\!{u'}_i^{\top  } \!{v}),
\end{align}
\noindent where $\{u_1,...,u_T\}$ are sampled from $p_d(u|v)$ and $\{u'_1, ...,{u'}_{k}^{\top}\}$ are sampled from $p_n(u|v)$. 

In this section, we only consider optimization of node $v$, which can be directly generalized to the whole embedding learning. More precisely, we equivalently consider $\*\theta = [\!{u_0}^\top \!{v}, ..., \!{u}_{N-1}^\top \!{v}]$ as the parameters to be optimized, where $\{u_0,...u_{N-1}\}$ are all the N nodes in the graph. 
Suppose the optimal parameter for $J^{(v)}_T$ and $J^{(v)}$ are $\*\theta_T$ and $\*\theta^*$ respectively. The gap between $\*\theta_T$ and $\*\theta^*$ is described as follows:
\begin{theorem}
    The random variable $\sqrt{T}(\*\theta_T - \*\theta^*)$ asymptotically converges to a distribution with zero mean vector and covariance matrix
    \begin{align}
        \text{Cov}\big(\sqrt{T}(\*\theta_T - \*\theta^*)\big) = \textbf{diag}(\*m)^{-1} - (1+1/k)\*1^\top\*1,
    \end{align}
\noindent where $\*m = \big[\frac{kp_d(u_0|v)p_n(u_0|v)}{p_d(u_0|v) + kp_n(u_0|v)}, ...,\frac{kp_d(u_{N-1}|v)p_n(u_{N-1}|v)}{p_d(u_{N-1}|v) + kp_n(u_{N-1}|v)}\big]^\top$ and $\*1 = [1, ...,1]^\top$.
\end{theorem}
\textbf{Proof\ } Our analysis is based on the Taylor expansion of $\nabla_\theta J_T^{(v)}(\*\theta)$ around $\*\theta^*$. As  $\*\theta_T$ is the minimizer of $J_T^{(v)}(\*\theta)$, we must have $\nabla_\theta J_T^{(v)}(\*\theta_T) = \*0$, which gives
\begin{align}
    \nabla_\theta J_T^{(v)}(\*\theta_T)  = \nabla J_T^{(v)}(\*\theta^*) + \nabla^2_\theta J_T^{(v)}(\*\theta^*)(\*\theta_T - \*\theta^*) + O(\|\*\theta_T - \*\theta^*\|^2)=\*0.
\end{align}
Up to terms of order $O(\|\*\theta^* - \*\theta_T\|^2)$, we have 
\begin{align}
    \sqrt{T}(\*\theta_T - \*\theta^*) = -\big(\nabla^2_{\theta} J_T^{(v)}(\*\theta^*)\big)^{-1} \sqrt{T} \nabla_{\theta}J_T^{(v)}(\*\theta^*).\label{break}
\end{align}

Next, we will analyze $-\big(\nabla^2_{\theta} J_T^{(v)}(\*\theta^*)\big)^{-1}$ and $\sqrt{T} \nabla_{\theta}J_T^{(v)}(\*\theta^*)$ respectively by the following lemmas.
\begin{lemma}
The negative Hessian matrix $-\nabla^2_{\theta} J_T^{(v)}(\*\theta^*)$ converges in probability to $\textbf{diag}(\*m)$.
\end{lemma}
\textbf{Proof\ } First, we calculate the gradient of $J_T^{(v)}(\*\theta)$ and Hessian matrix $\nabla_\theta^2 J^{(v)}_T(\*\theta)$.
Let $\*\theta_{u}$ be the parameter in $\*\theta$ for modeling $\!u^\top\!v$ and  $\*e_{(u)}$ be the one-hot vector which only has a 1 on this dimension. 
\begin{align}
    J^{(v)}_T(\*\theta) = &\frac{1}{T}\sum_{i=1}^T\log \sigma(\*\theta_{u_i}) + \frac{1}{T}\sum_{i=1}^{kT}\log \sigma(-\*\theta_{u'_i})\nonumber\\
    \nabla_\theta J^{(v)}_T(\*\theta) =& \frac{1}{T}\sum_{i=1}^T \big(1- \sigma(\*\theta_{u_i})\big)\*e_{(u_i)} - \frac{1}{T}\sum_{i=1}^{kT}\sigma(\*\theta_{u'_i})\*e_{(u'_i)}\label{gradient}\\
    \nabla_\theta^2 J^{(v)}_T(\*\theta) =& \frac{1}{T}\sum_{i=1}^T\{-\sigma(\*\theta_{u_i})\left(1-\sigma(\*\theta_{u_i})\right)\*e_{(u_i)}\*e_{(u_i)}^\top + \*0^\top\*0\} \nonumber\\
    &- \frac{1}{T}\sum_{i=1}^{kT}\{\sigma(\*\theta_{u'_i})\left(1-\sigma(\*\theta_{u'_i})\right)\*e_{(u'_i)}\*e_{(u'_i)}^\top + \*0^\top\*0\}.\nonumber
\end{align}
According to equation~\eqref{mterm}, $\sigma(\*\theta_{u_i})=\frac{p_d(u|v)}{p_d(u|v)+kp_n(u|v)}$ at $\*\theta = \*\theta^*$. Therefore, 
\begin{align}
    \lim\limits_{T\rightarrow +\infty}-\nabla_\theta^2 J^{(v)}_T(\*\theta^*) \xrightarrow{P}& \sum\limits_u \sigma(\*\theta_{u})\left(1-\sigma(\*\theta_{u})\right)\*e_{(u)}\*e_{(u)}^\top\nonumber\\ &\cdot(p_d(u)+kp_n(u))\nonumber\\
    =&\sum\limits_u\frac{kp_d(u|v)p_n(u|v)}{p_d(u|v) + kp_n(u|v)}\*e_{(u)}\*e_{(u)}^\top\nonumber\\
    =& \ \textbf{diag}(\*m).\label{lemma1}&\qed
\end{align}
\begin{lemma} The expectation and variance of $\nabla_\theta J_T^{(v)}(\*\theta^*)$ satisfy
\begin{align}
    \E [\nabla_\theta J_T^{(v)}(\*\theta^*)] &= \*0,\\
    \text{Var }[\nabla_\theta J_T^{(v)}(\*\theta^*)] &= \frac{1}{T}\big(\textbf{diag}(\*m) - (1+1/k)\*m\*m^\top\big).
\end{align}
\end{lemma}
\textbf{Proof\ } According to equation~\eqref{mterm}\eqref{gradient},
\begin{align*}
    \E [\nabla_\theta J_T^{(v)}(\*\theta^*)] &= \sum\limits_u [p_d(u)\big(1- \sigma(\*\theta_{u}^*)\big)\*e_{(u)} - kp_n(u)\sigma(\*\theta_{u}^*)\*e_{(u)}]=\*0,
\end{align*}
\begin{align}
    \text{Cov }[\nabla_\theta J_T^{(v)}(\*\theta^*)] =& \E [\nabla_\theta J_T^{(v)}(\*\theta^*)\nabla_\theta J_T^{(v)}(\*\theta^*)^\top]\nonumber\\
    =& \frac{1}{T}\E_{u\sim p_d(u|v)} \left(1-\sigma(\*\theta_{u}^*)\right)^2\*e_{(u)}\*e_{(u)}^\top\nonumber\\
    &+ \frac{k}{T}\E_{u\sim p_n(u|v)} \sigma(\*\theta_{u}^*)^2\*e_{(u)}\*e_{(u)}^\top\nonumber\\
    &+(\frac{T-1}{T} - 2 + 1 - \frac{1}{kT})\*m\*m^\top\nonumber\\
    =& \frac{1}{T}\big(\textbf{diag}(\*m) - (1+\frac{1}{k})\*m\*m^\top\big). \label{lemma2} &\qed
\end{align}

Let us continue to prove Theorem 2 by substituting~\eqref{lemma1}\eqref{lemma2} into equation \eqref{break} and derive the covariance of $\sqrt{T}(\*\theta_T - \*\theta^*)$.
\begin{align*}
    \text{Cov}\big(\sqrt{T}(\*\theta_T - \*\theta^*)\big)&= \E[\sqrt{T}(\*\theta_T - \*\theta^*)\ \sqrt{T}(\*\theta_T - \*\theta^*)^\top] \\
    &\approx T\textbf{diag}(\*m)^{-1}\text{Var }[\nabla_\theta J_T^{(v)}(\*\theta^*)]\big(\textbf{diag}(\*m)^{-1}\big)^\top\\
    &=\textbf{diag}(\*m)^{-1} - (1+1/k)\*1^\top\*1.
    \tag*{\qed}
\end{align*}
According to Theorem 2, the mean squared error, aka risk, of $\!u^\top\!v$ is \begin{align}\E \big[\|(\*\theta_T - \*\theta^*)_u\|^2\big] = \frac{1}{T}(\frac{1}{p_d(u|v)} - 1 + \frac{1}{kp_n(u|v)} - \frac{1}{k}).\label{deviation}\end{align}

\subsection{The Principle of Negative Sampling}\label{principle}
To determine an appropriate $p_n$ for a specific $p_d$, a trade-off might be needed to balance the rationality of the \textit{objective} (to what extent the optimal embedding fits for the downstream task) and the expected loss, also known as \textit{risk} (to what extent the trained embedding deviates from the optimum). 

A simple solution is to sample negative nodes positively but sub-linearly correlated to their positive sampling distribution, i.e. $p_n(u|v)\propto p_d(u|v)^\alpha, 0 < \alpha < 1$.

\textbf{(Monotonicity)} 
From the perspective of objective, if we have $p_d(u_i |v) > p_d(u_j | v)$, 
\begin{align*}
\!u^\top_i\!v &= \log p_d(u_i|v)- \alpha\log p_d(u_i | v) + c \\ &> (1-\alpha)\log p_d(u_j|v) + c = \!u^\top_j\!v,\end{align*}
where $c$ is a constant. The sizes of inner products is in accordance with positive distribution $p_d$, facilitating the task relying on relative sizes of $\!u^\top\!v$ for different $u$s, such as recommendation or link prediction. 
\label{monotonicity}

\textbf{(Accuracy)} From the perspective of risk, we mainly care about the scale of the expected loss. According to equation~\eqref{deviation}, the expected squared deviation is dominated by the smallest one in $p_d(u|v)$ and $p_n(u|v)$. If an intermediate node with large $p_d(u|v)$ is negatively sampled insufficiently (with small $p_n(u|v)$), the accurate optimization will be corrupted. But if $p_n(u|v)\propto p_d(u|v)^\alpha$, then
\begin{align*}\E \big[\|(\*\theta_T - \*\theta^*)_u\|^2\big] = \frac{1}{T}\big(\frac{1}{p_d(u|v)}(1+ \frac{p_d(u|v)^{1-\alpha}}{c}) -1- \frac{1}{k}\big).\end{align*}
The order of magnitude of deviation only negatively related to $p_d(u|v)$, meaning that with limited positive samples, the inner products for high-probability node pairs are estimated more accurate. This is an evident advantage over evenly negative sampling. 



\section{Method} 
\label{sec:method}

\begin{figure*}
    \centering 
    \setlength{\abovecaptionskip}{3pt}
    \setlength{\belowcaptionskip}{-2pt}
    \includegraphics[width=\textwidth]{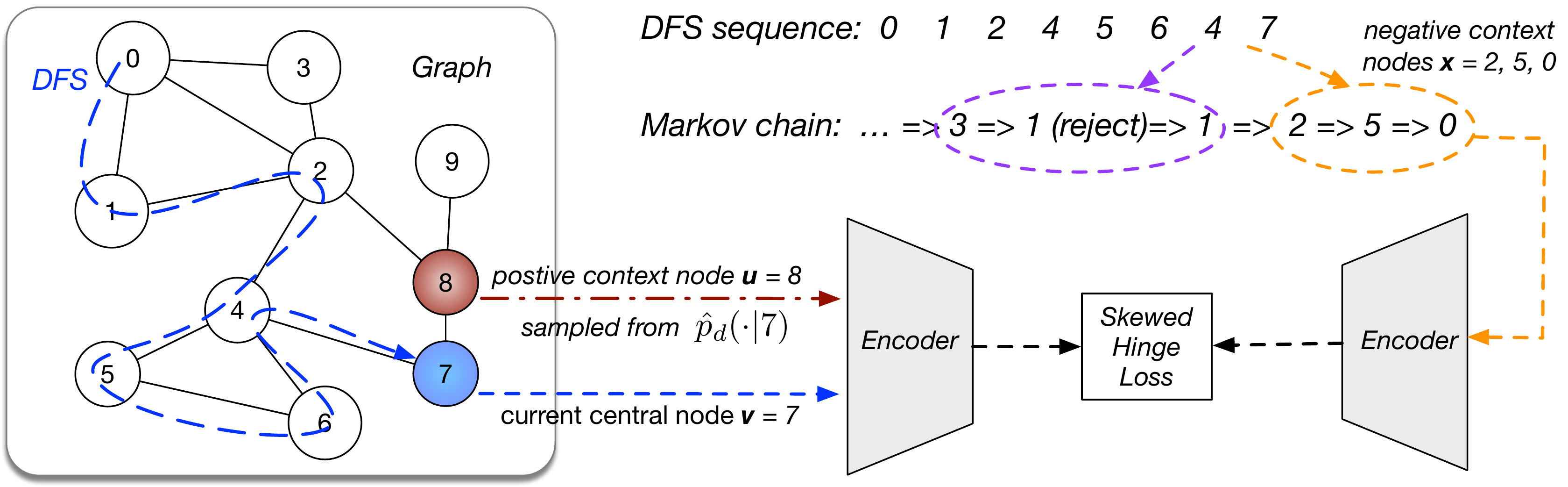}
    \caption{A running example of \model. The central nodes are traversed by DFS, each samples three negative context nodes using Metropolis-Hastings algorithm by proceeding with the Markov chain.}
    \label{fig:model}
    \end{figure*}

\begin{algorithm}[t]
\caption{The training process with \model}\label{algo}
\KwIn{
DFS sequence $T=[v_1,...,v_{|T|}]$,\\ Encoder ${E}_\theta$,  Positive Sampling Distribution $\hat{p_d}$, \\Proposal Distribution $q$,  Number of Negative Samples $k$.
}
\Repeat{Early Stopping or Convergence}{
    Initialize current negative node $x$ at random\\
    \For{each node $v$ in DFS sequence $T$}{
        Sample $1$ positive sample $u$ from $\hat{p_d}(u|v)$\\
        Initialize loss $\&L = 0$\\
        \textcolor{comment}{//Negative Sampling via Metropolis-Hastings}\\
        \For{$i=1$ to $k$}{
         Sample a node $y$ from $q(y|x)$\\
         Generate a uniform random number $r\in [0,1]$\\
         \If{$r < \text{min}\big(1, \frac{(E_\theta(v)\cdot E_\theta(y))^{\alpha}}{(E_\theta(v)\cdot E_\theta(x))^{\alpha}} \frac{q(x|y)}{q(y|x)}\big)$}{ 
                $x= y$\\
         }
         $\&L = \&L + \text{max}\big(0,E_{\theta}(v)\cdot E_{\theta}(x) - E_{\theta}(u)\cdot E_{\theta}(v)+ \gamma\big)$\\
         Update $\theta$ by descending the gradients  $\nabla_\theta\&L$\\
        }
    }
}
\end{algorithm}
\subsection{The Self-contrast Approximation}
Although we deduced that $p_n(u|v)\propto p_d(u|v)^\alpha$, the real $p_d$ is unknown and its approximation $\hat{p_d}$ is often implicitly defined. -- How can the principle help negative sampling?

We therefore propose a \emph{self-contrast approximation}, replacing $p_d$ by inner products based on the current encoder, i.e.
\[p_n(u|v)\propto p_d(u|v)^\alpha \approx \frac{\big(E_\theta(u)\cdot E_\theta(v)\big)^\alpha}{\sum_{u'\in U}\big(E_\theta(u')\cdot E_\theta(v)\big)^\alpha}.\]
The resulting form is similar to a technique in RotatE~\cite{sun2019rotate}, and very time-consuming. Each sampling requires $O(n)$ time, making it impossible for middle- or large-scale graphs. Our \smodel solves this problem by our adapted version of Metropolis-Hastings algorithm.
\subsection{The Metropolis-Hastings Algorithm}
As one of the most distinguished Markov chain Monte Carlo (MCMC) methods, the Metropolis-Hastings algorithm~\cite{metropolis1953equation} is designed for obtaining a sequence of random samples from unnormalized distributions. Suppose we want to sample from distribution $\pi(x)$ but only know $\tilde{\pi}(x)\propto \pi(x)$. Meanwhile the computation of normalizer $Z=\sum_x\tilde{\pi}(x)$ is unaffordable. The Metropolis-Hastings algorithm constructs a Markov chain $\{X(t)\}$ that is ergodic and stationary with respect to $\pi$ —- meaning that, $X(t)\sim \pi(x), t\rightarrow \infty$. 

The transition in the Markov chain has two steps:\\(1) Generate $y\sim q(y|X(t))$ where the \emph{proposal distribution} $q$ is arbitrary chosen as long as positive everywhere.\\(2) Pick $y$ as $X(t+1)$ at the probability $\min \left\{\frac{\tilde{\pi}(y)}{\tilde{\pi}(X(t))} \frac{q(X(t) | y)}{q(y | X(t))}, 1\right\}$, or $X(t+1)\leftarrow X(t)$. See tutorial~\cite{chib1995understanding} for more details.
\subsection{Markov chain Negative Sampling}
The main idea for \smodel is to apply the Metropolis-Hastings algorithm for each node $v$ with $\tilde{\pi}(u|v)=\big(E_\theta(u)\cdot E_\theta(v)\big)^\alpha$. Then we are sampling from the self-contrast approximated distribution. However, a notorious disadvantage for the Metropolis-Hastings is a relatively long $burn-in$ period, i.e. the distribution of $X(t)
\approx\pi(x)$ only when $t$ is large, which heavily affects its efficiency. 

Fortunately, the connatural property exists for graphs that nearby nodes usually share similar $p_d(\cdot|v)$, as evident from triadic closure in social network~\cite{huang2014mining},  collaborative filtering in recommendation~\cite{linden2003amazon}, etc. Since nearby nodes have similar target negative sampling distribution, the Markov chain for one node is likely to still work well for its neighbors. Therefore, a neat solution is to traverse the graph by Depth First Search (DFS, See Appendix~\ref{app:dfs}) and keep generating negative samples from the last node's Markov chain (Figure~\ref{fig:model}).

The proposal distribution also influences convergence rate. Our $q(y|x)$ is defined by mixing uniform sampling and sampling from the nearest $k$ nodes with $\frac{1}{2}$ probability each. 
The hinge loss pulls positive node pairs closer and pushes negative node pairs away until they are beyond a pre-defined margin. Thus, we substitute the binary cross-entropy loss as a $\gamma$-skewed hinge loss,
\begin{equation}
\&L = \text{max}\big(0,E_{\theta}(v)\cdot E_{\theta}(x) - E_{\theta}(u)\cdot E_{\theta}(v)+ \gamma\big),\label{eq:margin}
\end{equation}
where $(u,v)$ is a positive node pair and $(x, v)$ is negative. $\gamma$ is a hyperparameter. The \smodel is summarized in Algorithm~\ref{algo}.

\section{Experiments} \label{sec:experiment}

To demonstrate the efficacy of \model, we conduct a comprehensive suite of experiments on 3 representative tasks, 3 graph representation learning algorithms and 5 datasets, a total of 19 experimental settings, under all of which \smodel consistently achieves significant improvements over 8 negative sampling baselines collected from previous papers~\cite{pinsage,deepwalk} and other relevant topics~\cite{dns,wang2017irgan,cai2017kbgan,weston2011wsabie}.
\subsection{Experimental Setup}
\subsubsection{\textbf{Tasks and Dataset.}}
The statistics of tasks and datasets are shown in Table~\ref{dataset}. Introductions about the datasets are in Appendix~\ref{app:dataset}.

\begin{table}[hbt]
\setlength{\abovecaptionskip}{0.3cm}
\setlength{\belowcaptionskip}{-0.3cm}
\centering
\caption{Statistics of the tasks and datasets.}
\label{dataset}
\renewcommand{\arraystretch}{1.15}
\setlength{\tabcolsep}{1mm}{
\begin{tabular}{c|c|c|c|c}  
\toprule
\multirow{5}{*}{}   Task & Dataset    & Nodes & Edges  & Classes \\
\midrule
\multirow{3}{*}{Recommendation}
        & MovieLens & 2,625   &  100,000   & / \\
        & Amazon    & 255,404 &  1,689,188 & / \\
        & Alibaba  & 159,633 &  907,470   & / \\  
\hline
\multirow{1}{*}{Link Prediction}
        & Arxiv & 5,242 & 28,980 & /\\
\hline
\multirow{1}{*}{Node Classification}
        & BlogCatalog & 10,312 & 333,983& 39  \\
    \bottomrule
    \end{tabular}}
    \vspace*{-0.5cm}
\end{table}

\subsubsection{\textbf{Encoders.}} To verify the adaptiveness of \smodel to different genres of graph representation learning algorithms, we utilize an embedding lookup table or a variant of GNN as encoder for experiments, including DeepWalk, GCN and GraphSAGE. The detailed algorithms are shown in Appendix~\ref{app:encoders}.

\begin{small}
    \begin{table*}[hbpt]
    \centering
    \renewcommand{\arraystretch}{1.15}
    \setlength{\tabcolsep}{2mm}{
    \begin{tabular}{c|c|ccc|cc|cc}  
    \toprule
    \multirow{5}{*}{} &  & \multicolumn{3}{c|}{MovieLens} & \multicolumn{2}{c|}{Amazon} & \multicolumn{2}{c}{Alibaba} \\
         &    & DeepWalk & GCN &  GraphSAGE & DeepWalk & GraphSAGE  & DeepWalk & GraphSAGE \\
    \midrule
    \multirow{9}{*}{$MRR$}
             & $Deg^{0.75}$  &  0.025$\pm$.001   & 0.062$\pm$.001   & 0.063$\pm$.001    & 0.041$\pm$.001   & 0.057$\pm$.001  & 0.037$\pm$.001   & 0.064$\pm$.001      \\ 
             & WRMF & 0.022$\pm$.001    &0.038$\pm$.001     & 0.040$\pm$.001    & 0.034$\pm$.001    & 0.043$\pm$.001  & 0.036$\pm$.001   & 0.057$\pm$.002    \\
             & RNS  &  0.031$\pm$.001   & 0.082$\pm$.002   &0.079$\pm$.001    & 0.046$\pm$.003   & 0.079$\pm$.003  & 0.035$\pm$.001  &  0.078$\pm$.003     \\
             & PinSAGE       &  0.036$\pm$.001    & 0.091$\pm$.002 & 0.090$\pm$.002     & 0.057$\pm$.004   & 0.080$\pm$.001  & 0.054$\pm$.001    & 0.081$\pm$.001       \\
             & WARP     & 0.041$\pm$.003    &0.114$\pm$.003   & 0.111$\pm$.003    & 0.061$\pm$.001   & 0.098$\pm$.002  & 0.067$\pm$.001  & 0.106$\pm$.001   \\
             & DNS  & 0.040$\pm$.003    & 0.113$\pm$.003   & 0.115$\pm$.003   & 0.063$\pm$.001 & 0.101$\pm$.003  & 0.067$\pm$.001   &0.090$\pm$.002  \\
             & IRGAN & 0.047$\pm$.002  &0.111$\pm$.002   & 0.101$\pm$.002   &0.059$\pm$.001   &0.091$\pm$.001  & 0.061$\pm$.001  & 0.083$\pm$.001   \\
             & KBGAN    & 0.049$\pm$.001 &0.114$\pm$.003    &0.100 $\pm$.001   & 0.060$\pm$.001   &0.089$\pm$.001  & 0.065$\pm$.001 & 0.087$\pm$.002     \\
             & \model        & \textbf{0.053$\pm$.001} & \textbf{0.122$\pm$.004} & \textbf{0.114$\pm$.001}  &\textbf{0.065$\pm$.001}    & \textbf{0.108$\pm$.001} & \textbf{0.070$\pm$.001} & \textbf{0.116$\pm$.001}       \\
    \hline
    
    \multirow{9}{*}{$Hits@30$}
             & $Deg^{0.75}$  &  0.115$\pm$.002   &0.270$\pm$.002    &0.270$\pm$.001   & 0.161$\pm$.003   & 0.238$\pm$.002  & $0.138\pm$.003  & 0.249$\pm$.004    \\ 
             & WRMF & 0.110$\pm$.003    & 0.187$\pm$.002    &0.181$\pm$.002   & 0.139$\pm$.002    & 0.188$\pm$.001  & 0.121$\pm$.003  & 0.227$\pm$.004    \\
             & RNS  &  0.143$\pm$.004  & 0.362$\pm$.004    &0.356$\pm$.001   & 0.171$\pm$.004    & 0.317$\pm$.004  & 0.132$\pm$.004  &  0.302$\pm$.005     \\
             & PinSAGE  &  0.158$\pm$.003  & 0.379$\pm$.005    & 0.383$\pm$.005   & 0.176$\pm$.004  &0.333$\pm$.005   & 0.146$\pm$.003    &0.312$\pm$.005  \\
             & WARP     & 0.164$\pm$.005   &0.406$\pm$.002    &0.404$\pm$.005   & 0.181$\pm$.004    & 0.340$\pm$.004  & 0.178$\pm$.004  & 0.342$\pm$.004   \\
             & DNS& 0.166$\pm$.005    & 0.404$\pm$.006    &0.410$\pm$.006   & 0.182$\pm$.003  & 0.358$\pm$.004  & 0.186$\pm$.005  &0.336$\pm$.004   \\
             & IRGAN & 0.207$\pm$.002    & 0.415$\pm$.004    &0.408$\pm$.004   & 0.183$\pm$.004   &0.342$\pm$.003  & 0.175$\pm$.003 & 0.320$\pm$.002   \\
             & KBGAN & 0.198$\pm$.003   & 0.420$\pm$.003    &0.401$\pm$.005   & 0.181$\pm$.003   & 0.347$\pm$.003  & 0.181$\pm$.003  & 0.331$\pm$.004     \\
            & \model        & \textbf{0.230$\pm$.003} & \textbf{0.426 $\pm$.005} & \textbf{0.413$\pm$.003}  &\textbf{0.207$\pm$.003}  & \textbf{0.386$\pm$.004} & \textbf{0.201$\pm$.003} & \textbf{0.387$\pm$.002}       \\
        \bottomrule
        \end{tabular}} 
    \caption{ Recommendation Results of \smodel with various encoders on three datasets. GCN is not evaluted on Amazon and Alibaba datasets due to its limited scalability. Similar trends hold for Hits@5 and Hits@10.}
    \label{result}
    \vspace*{-0.5cm}
    \end{table*}
\end{small}
\subsubsection{\textbf{Baselines.}}

Baselines in our experiments generally fall into the following categories. The detailed description of each baseline is provided in Appendix~\ref{app:baselines}.
\begin{itemize}
\item \textbf{Degree-based Negative Sampling.} The compared methods include Power of Degree \cite{word2vec}, RNS \cite{caselles2018word2vec} and WRMF \cite{hu2008collaborative}, which can be summarized as $p_n(v) \propto deg(v)^\beta$.
These strategies are static, inconsiderate to the personalization of nodes.

\item \textbf{Hard-samples Negative Sampling.} Based on the observation that ``hard negative samples'' -- the nodes with high positive probabilities -- helps recommendation, methods are proposed to mine the hard negative samples by rejection (WARP~\cite{zhao2015improving}), sampling-max (DNS~\cite{dns}) and personalized PageRank (PinSAGE~\cite{pinsage}).

\item \textbf{GAN-based Negative Sampling.} IRGAN~\cite{wang2017irgan} trains generative adversarial nets (GANs) to adaptively generate hard negative samples. KBGAN~\cite{cai2017kbgan} employs a sampling-based adaptation of IRGAN on knowledge base completion.
\end{itemize}
\subsection{Results}
In this section, we demonstrate the results in the 19 settings. Note that all values and standard deviations reported in the tables are from ten-fold cross validation. We also leverage paired t-tests~\cite{hsu2007paired} to verify whether \smodel is significantly better than the strongest baseline. The tests in the 19 settings give a max \textit{p-value} ({Amazon, GraphSAGE}) of {0.0005} $\ll 0.01$, quantitatively proving the significance of our improvements. 

\subsubsection{\textbf{Recommendation}}
Recommendation is the most important technology in many E-commerce platforms, which evolved from collaborative filtering to graph-based models.
Graph-based recommender systems represent all users and items by embeddings, and recommend items with largest inner products for a given user.

In this paper, $Hits@k$ \cite{hit1} and mean reciprocal ranking ($MRR$) \cite{mrr} serve as evaluation methodologies, whose definitions can be found in Appendix~\ref{app:metrics}. Moreover, we only use observed positive pairs to train node embeddings, which can clearly demonstrate that the improvement of performance comes from the proposed negative sampling. We summarize the performance of \smodel as well as the baselines in Table~\ref{result}. 

The results show that Hard-samples negative samplings generally surpass degree-based strategies 
with $5\% \sim 40\%$ performance leaps in MRR.
GAN-based negative sampling strategies perform even better, owing to the evolving generator mining hard negative samples more accurately. In the light of our theory, the proposed \smodel accomplishes significant gains of $2\%\sim13\%$ over the best baselines.


\subsubsection{\textbf{Link Prediction}}
Link prediction aims to predict the occurrence of links in graphs, more specifically, to judge whether a node pair is a masked true edge or unlinked pair based on nodes' representations. 
Details about data split and evaluation are in Appendix~\ref{app:metrics}.

The results for link prediction are given in Table~\ref{tab:pred}. \smodel outperforms all baselines with various graph representation learning methods on the Arxiv dataset. 
An interesting phenomenon about degree-based strategies is that both WRMF and commonly-used $Deg^{0.75}$ outperform uniform RNS sampling, completely adverse to the results on recommendation datasets. The results can be verified by examining the results in previous papers~\cite{word2vec, zhang2018gneg}, reflecting the different network characteristics of user-item graphs and citation networks.

\begin{table}[hbt]
\centering
\renewcommand{\arraystretch}{1.15}
\setlength{\tabcolsep}{1mm}{
\begin{tabular}{c|c|ccc}  
\toprule
     & Algorithm   & DeepWalk & GCN &  GraphSAGE \\
\midrule
\multirow{10}{*}{AUC}
         & $Deg^{0.75}$  &  64.6$\pm$0.1  & 79.6$\pm$0.4  & 78.9$\pm$0.4 \\
         & WRMF    & 65.3$\pm$0.1    & 80.3$\pm$0.4    & 79.1$\pm$0.2   \\
         & RNS     &  62.2$\pm$0.2   & 74.3$\pm$0.5    & 74.7$\pm$0.5     \\
         & PinSAGE &  67.2$\pm$0.4   & 80.4$\pm$0.3    & 80.1$\pm$0.4     \\
         & WARP    & 70.5$\pm$0.3    & 81.6$\pm$0.3    & 82.7$\pm$0.4    \\
         & DNS     & 70.4$\pm$0.3    & 81.5$\pm$0.3    & 82.6$\pm$0.4   \\
         & IRGAN   & 71.1$\pm$0.2  & 82.0$\pm$0.4  & 82.2$\pm$0.3  \\
         & KBGAN   & 71.6$\pm$0.3  & 81.7$\pm$0.3  & 82.1$\pm$0.3   \\
         & \model       & \textbf{73.1$\pm$0.4}  & \textbf{82.6$\pm$0.4}   & \textbf{83.5$\pm$0.5}   \\
    \bottomrule
    \end{tabular}}
\caption{The results of link prediction with different negative sampling strategies on the Arxiv dataset.}\label{tab:pred}
\label{result_link_prediction}
\vspace*{-0.6cm}
\end{table}

\subsubsection{\textbf{Node Classification}}

Multi-label node classification is a usual task to assess graph representation algorithms. The one-vs-rest logistic regression classifier\cite{fan2008liblinear} is trained in a supervised way with graph representations as the features of nodes. In this experiment, we keep the default setting as the DeepWalk’s original paper for fair comparison on the BlogCatalog dataset.

Table 4 summarizes the Micro-F1 result on the BlogCatalog dataset. \smodel stably outperforms all baselines regardless of the training set ratio $T_R$. The trends are similar for Macro-F1, which is omitted due to the space limitation.
\begin{small}
\begin{table}[hbt]
\centering
\renewcommand{\arraystretch}{1.15}
\setlength{\tabcolsep}{0.8mm}{
\begin{tabular}{c|c|ccc|ccc|ccc}  
\toprule
\multirow{2}{*}{} & Algorithm & \multicolumn{3}{c|}{DeepWalk} & \multicolumn{3}{c|}{GCN} & \multicolumn{3}{c}{GraphSAGE} \\
\hline
     & $T_R(\%)$ & 10  &50  & 90  & 10 & 50 & 90 & 10 & 50 & 90  \\
\midrule
\multirowcell{10}{Micro\\-F1}
         & $Deg^{0.75}$  &31.6 & 36.6 & 39.1   &36.1 & 41.8 & 44.6  &35.9 & 42.1 & 44.0 \\
         & WRMF         &30.9 &35.8 & 37.5 & 34.2 &41.4 & 43.3  & 34.4 & 41.0 & 43.1    \\
         & RNS       & 29.8 & 34.1 & 36.0 & 33.4 & 40.5 & 42.3 &33.5& 39.6& 41.6    \\
         & PinSAGE   & 32.0 & 37.4 & 40.1  & 37.2 & 43.2 & 45.7  & 36.9 & 43.2 & 45.1     \\
         & WARP  & 35.1 & 40.3 & 42.1  & 39.9 & 45.8 & 47.7  & 40.1 & 45.5 & 47.5         \\
         & DNS   & 35.2 & 40.4 & 42.5  & 40.4 & 46.0& 48.6  & 40.5 & 46.3 & 48.5         \\
         & IRGAN  & 34.3 & 39.6 & 41.8   & 39.1 & 45.2 & 47.9  & 38.9 & 45.0 & 47.6        \\
         & KBGAN  & 34.6 & 40.0 & 42.3  & 39.5 & 45.5 & 48.3  & 39.6 & 45.3 & 48.5        \\
         & \model      &\textbf{36.1} &\textbf{41.2} &\textbf{43.3} &\textbf{41.7} &\textbf{47.3} &\textbf{49.9} &\textbf{41.6} &\textbf{47.5} &\textbf{50.1} \\
    \bottomrule
    \end{tabular}}
\caption{Micro-F1 scores for multi-label classification on the BlogCatalog dataset. Similar trends hold for Macro-F1 scores.}
\label{result_link_prediction}
\vspace*{-0.85cm}
\end{table}
\end{small}

\subsection{Analysis} 
\begin{figure}[htbp]
  \centering
  \setlength{\abovecaptionskip}{3pt}
  \setlength{\belowcaptionskip}{-3pt}
  \includegraphics[width=\linewidth]{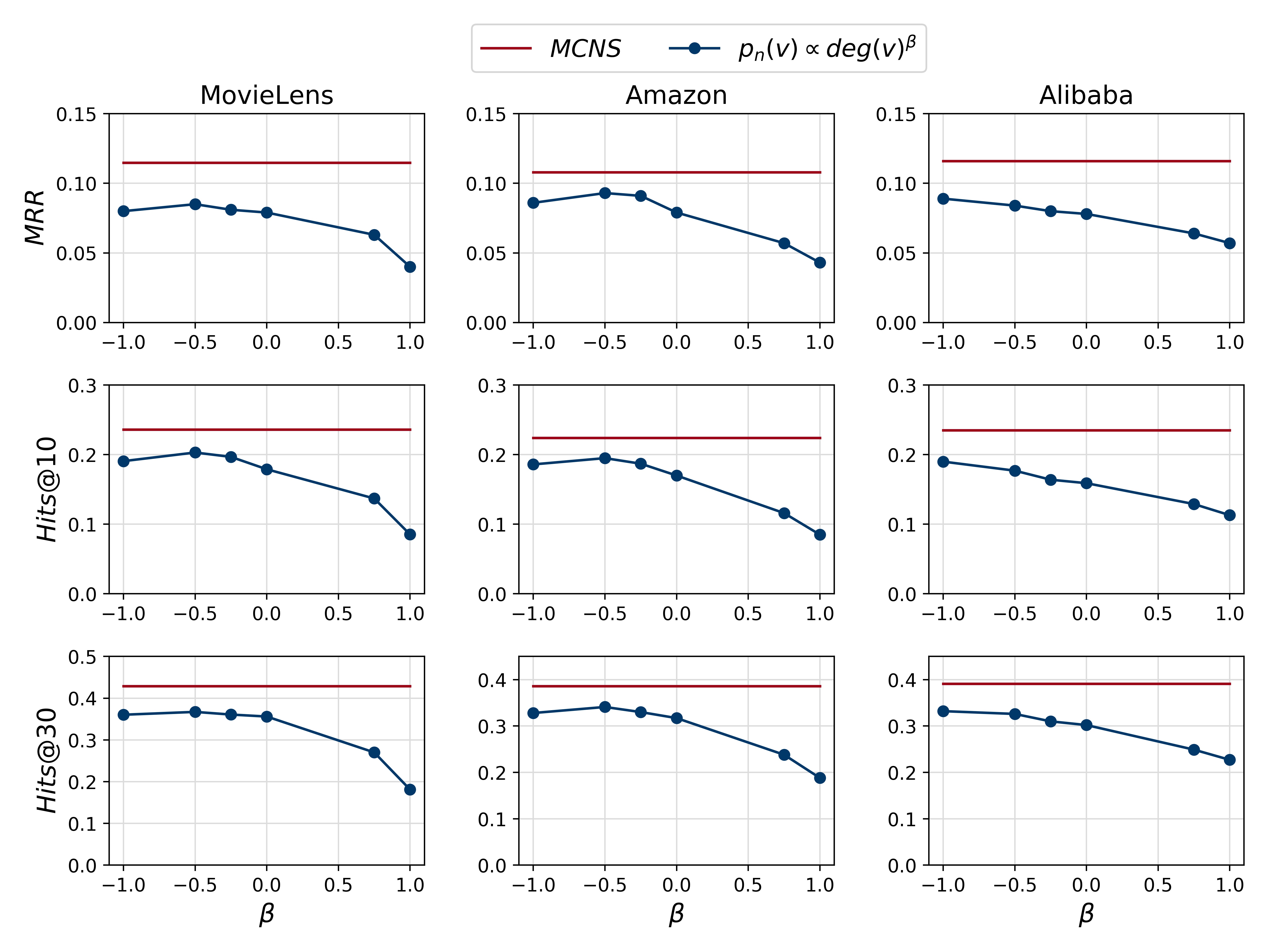}
  \caption{Comparison between Power of Degree and \model. Each red line represents the performance of \smodel in this setting and the blue curves are performances of Power of Degree with various $\beta$. }
  \label{degree}
\end{figure}

\noindent \textbf{Comparison with Power of Degree.} To thoroughly investigate the potential of degree-based strategies $p_n(v)\propto deg(v)^\beta$, a series of experiments are conducted by varying $\beta$ from -1 to 1 with GraphSAGE. Figure~\ref{degree} shows that the performance of degree-based is consistently  lower than that of \model, suggesting that \smodel learns a better negative distribution beyond the expressiveness of degree-based strategies. Moreover, 
the best $\beta$ varies between datasets and seems not easy to determine before experiments, while \smodel naturally adapts to different datasets. 

\begin{figure}
\centering 
\setlength{\abovecaptionskip}{3pt}
\setlength{\belowcaptionskip}{-1pt}
\includegraphics[width=\linewidth]{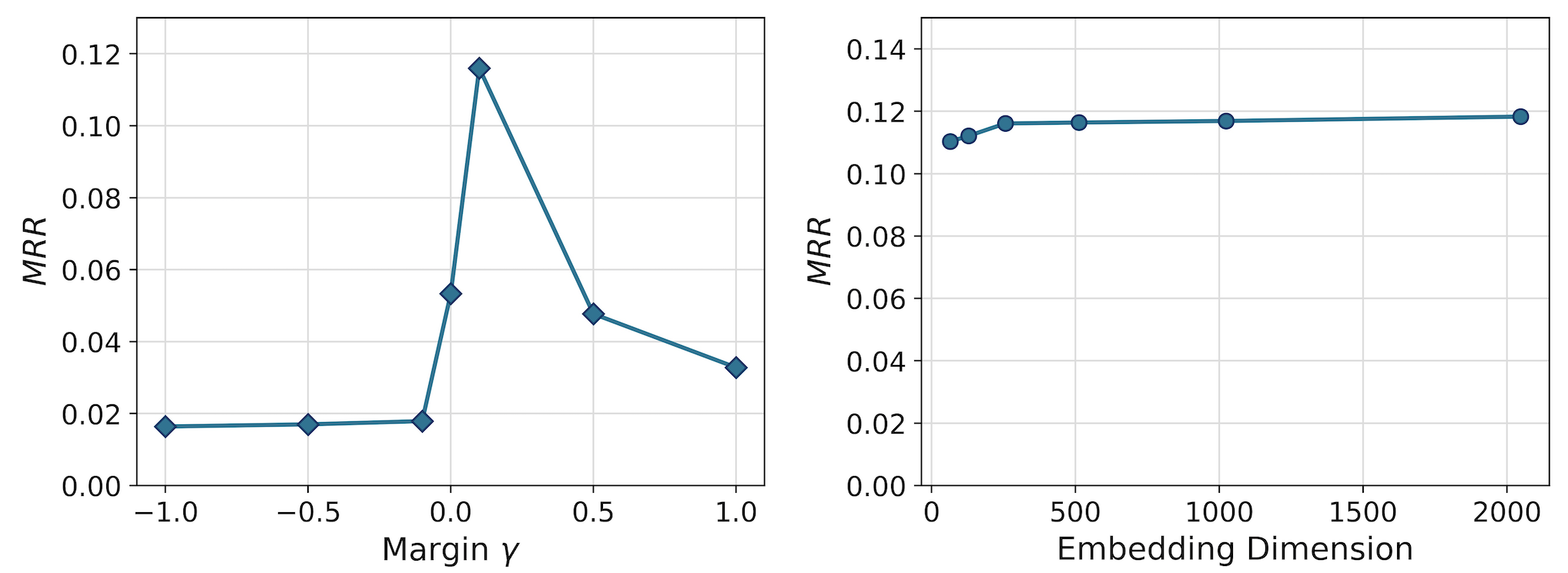}
\caption{The performance of \model \ on the Alibaba by varying $\gamma$ and embedding dimension. }
\label{params}
\vspace{-0.5cm}
\end{figure}

\noindent \textbf{Parameter Analysis.}
To test the robustness of \smodel quantitively, we visualize the MRR curves of \smodel by varying two most important hyperparameters, embedding dimension and margin $\gamma$. The results with GraphSAGE on Alibaba dataset are shown in Figure~\ref{params}.

The skewed hinge loss in Equation~\eqref{eq:margin} aims to keep at least $\gamma$ distance between the inner product of the positive pair and that of the negative pair. 
$\gamma$ controls the distance between positive-negative samples and must be positive in principle. Correspondingly, Figure~\ref{params} shows that the hinge loss begins to take effect when $\gamma \ge 0$ and reaches its optimum at $\gamma\approx 0.1$, which is a reasonable boundary. 
As the performance increases slowly with a large embedding dimension, we set it $512$ due to the trade-off between the performance and time consumption.

\begin{figure}[htbp]
  \centering
  \setlength{\abovecaptionskip}{3pt}
  \setlength{\belowcaptionskip}{-3pt}
  \includegraphics[width=\linewidth]{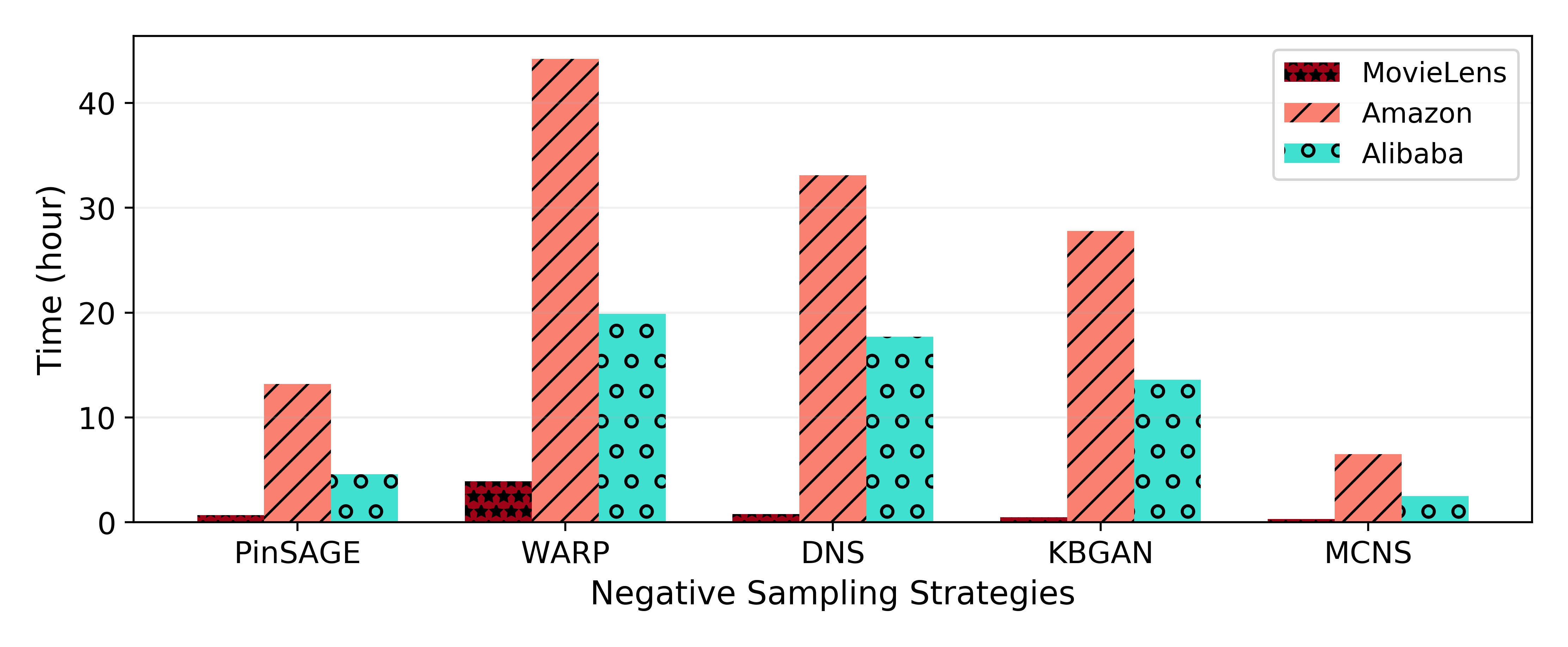}
  \caption{Runtime comparisons for different negative sampling strategies with GraphSAGE encoder on the recommendation datasets.}
  \label{fig:time}
  \vspace{-0.2cm}
\end{figure}

\noindent \textbf{Efficiency Comparison.} To compare the efficiency of different negative sampling methods, we report the runtime of \smodel and  hard-samples or GAN-based strategies (PinSAGE, WARP, DNS, KBGAN) with GraphSAGE encoder in recommendation task in Figure~\ref{fig:time}. We keep the same batch size and number of epochs for a fair comparison. WARP might need a large number of tries before finding negative samples, leading to a long running time. PinSAGE utilizes PageRank to generate ``hard negative items'', which increases running time compared with uniformly sampling. DNS and KBGAN both first sample a batch of candidates and then select negative samples from them, but the latter uses a lighter generator. \model \ requires 17 mins to complete ten-epochs training while the fastest baseline KBGAN is at least 2$\times$ slower on the MovieLens dataset.

\subsection{Experiments for Further Understanding}
In section~\ref{sec:theory}, we have analyzed the criterion about negative sampling from the perspective of {objective and risk}. 
However, doubts might arise about (1) whether sampling more negative samples is always helpful,
(2) why our conclusion contradicts with the intuition ``positively sampling nearby nodes and negatively sampling faraway nodes ''. 

\begin{figure}[htbp]
    \centering 
    \setlength{\abovecaptionskip}{3pt}
    \setlength{\belowcaptionskip}{-3pt}
    \includegraphics[width=\linewidth]{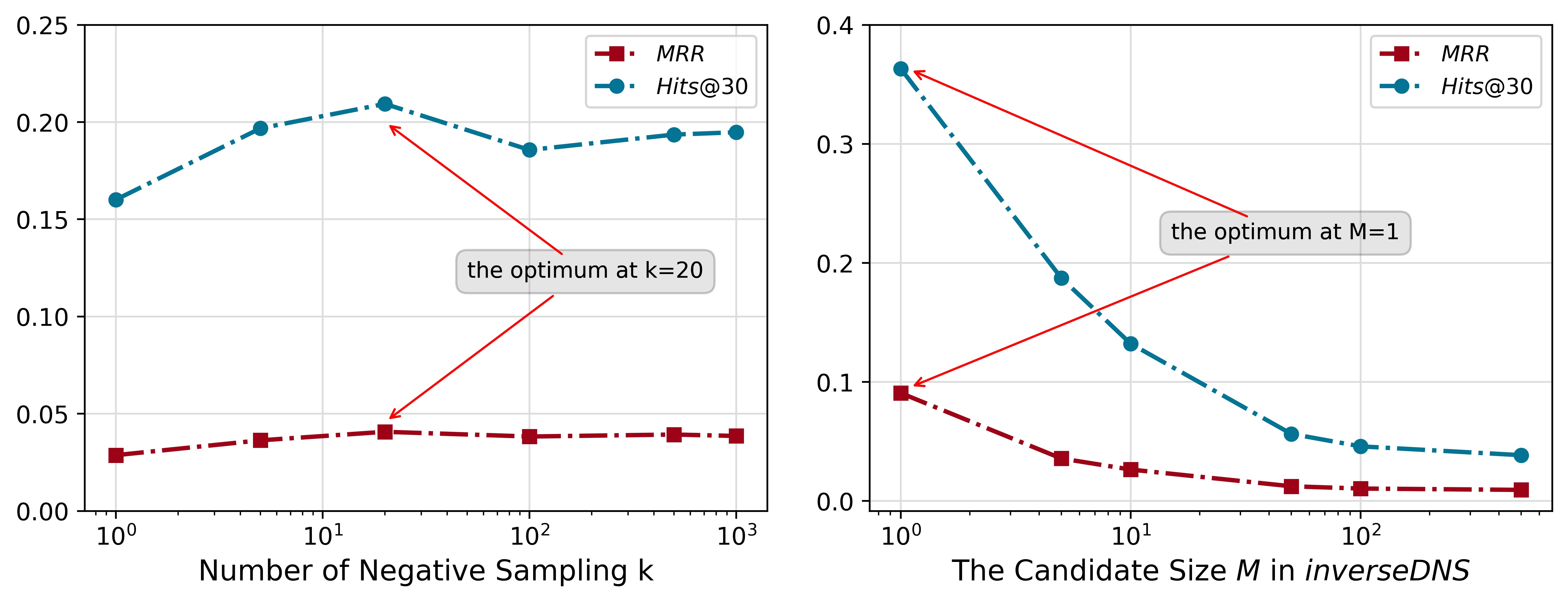}
    \caption{The results of verification experiments. (Left) The effect of increasing the number of negative sampling $k$ in $Deg^{0.75}$. (Right) The effect of sampling more \textit{distant nodes}.}
    \label{verification}
    \vspace*{-0.2cm}
\end{figure}

To further deepen the understanding of negative sampling, we design two extended experiments on MovieLens with GraphSAGE encoder to verify our theory and present the results in Figure~\ref{verification}.

If we sample more negative samples, the performance increases with subsequently decrease.
According to Equation~\eqref{deviation}, sampling more negative samples always decrease the risk, leading to an improvement in performance at first. 
However, performance begins to decrease after reaching the optimum, because extra bias is added to the objective by the increase of $k$ according to Equation~\eqref{obj}. The bias impedes the optimization of the objective from zero-centered initial embeddings. 

The second experiment in Figure~\ref{verification} shows the disastrous consequences of negatively sampling more \textit{distant nodes}, the nodes with small $p_d(u)$. To test the performances with varying degrees of the mismatching of $p_d$ and $p_n$,  we design a strategy, \textit{inverseDNS}, by selecting the one scored lowest in the candidate items. In this way, not only the negative sampling probabilities of the items are sampled negatively correlated to $p_d$, but also we can control the degree by varying candidate size $M$. Both MRR and $Hits@k$ go down as $M$ increases, hence verifies our theory.



\section{Related Work} \label{sec:related}
\textbf{Graph Representation Learning.}
The mainstream of graph representation learning on graphs diverges into two main topics: traditional network embedding and GNNs. Traditional network embedding cares more about the distribution of positive node pairs. Inspired by the skip-gram model \cite{word2vec}, DeepWalk \cite{deepwalk} learns embeddings via sampling ``context'' nodes for each vertex with random walks, and maximize the log-likelihood of observed context nodes for the given vertex. 
LINE \cite{line} and node2vec \cite{node2vec} extend DeepWalk with various positive distributions.
GNNs are deep learning based methods that generalize convolution operation to graph data. 
\citep{gcn} design GCNs by approximating localized 1-order spectral convolution.
For scalability, GraphSAGE  \cite{graphsage} employs neighbor sampling to alleviate the receptive field expansion. FastGCN \cite{fastgcn} adopts importance sampling in each layer. 

\textbf{Negative Sampling.} Negative sampling is firstly proposed to speed up skip-gram training in word2vec \cite{word2vec}. Word embedding models sample negative samples according to its word frequency distribution proportional to the 3/4 power. Most later works  on network embedding \cite{node2vec, line} follow this setting. In addition, negative sampling has been studied extensively in recommendation. Bayesian Personalized Ranking~\cite{rendle2009bpr} proposes uniform sampling for negative samples. Then, dynamic negative sampling (DNS) \cite{dns} is proposed to sample informative negative samples based on current prediction scores. 
Besides, the PinSAGE \cite{pinsage} adds ``hard negative items'' to obtain fine enough ``resolution'' for recommender system.
The negative samples are sampled uniformly with the rejection in WARP \cite{weston2011wsabie}. 
Recently, GAN-based negative sampling method has also been adopted to train better node representations for information retrieval \cite{wang2017irgan}, recommendation \cite{wang2018neural}, word embeddings \cite{bose2018adversarial}, knowledge graph embeddings \cite{cai2017kbgan}. 
Another lines of research focus on exploiting  sampled softmax~\citep{sampledsoftmax_raw} or its debiasing variant~\citep{zhou2020contrastive} for extremely large scale link prediction tasks.

\section{Conclusion}  \label{sec:conclusion}
In this paper, we study the effect of negative sampling, a practical approach adopted in the literature of graph representation learning. 
Different from the existing works that decide a proper distribution for negative sampling in heuristics, we theoretically analyze the objective and risk of the negative sampling approach and conclude that the negative sampling distribution should be positively but sub-linearly correlated to their positive sampling distribution.
Motivated by the theoretical results, we propose \model, approximating the ideal distribution by self-contrast and accelerating sampling by Metropolis-Hastings. 
Extensive experiments show that \smodel outperforms 8 negative sampling strategies, regardless of the underlying graph representation learning methods.

\section*{Acknowledgements}
The work is supported by the NSFC for Distinguished Young Scholar (61825602), NSFC (61836013), and a research fund supported by Alibaba Group.

%
\bibliographystyle{ACM-Reference-Format}
\bibliography{main}


\begin{thebibliography}{46}


\ifx \showCODEN    \undefined \def \showCODEN     #1{\unskip}     \fi
\ifx \showDOI      \undefined \def \showDOI       #1{#1}\fi
\ifx \showISBNx    \undefined \def \showISBNx     #1{\unskip}     \fi
\ifx \showISBNxiii \undefined \def \showISBNxiii  #1{\unskip}     \fi
\ifx \showISSN     \undefined \def \showISSN      #1{\unskip}     \fi
\ifx \showLCCN     \undefined \def \showLCCN      #1{\unskip}     \fi
\ifx \shownote     \undefined \def \shownote      #1{#1}          \fi
\ifx \showarticletitle \undefined \def \showarticletitle #1{#1}   \fi
\ifx \showURL      \undefined \def \showURL       {\relax}        \fi
\providecommand\bibfield[2]{#2}
\providecommand\bibinfo[2]{#2}
\providecommand\natexlab[1]{#1}
\providecommand\showeprint[2][]{arXiv:#2}

\bibitem[\protect\citeauthoryear{Bengio and Sen{\'e}cal}{Bengio and
  Sen{\'e}cal}{2008}]%
        {sampledsoftmax_raw}
\bibfield{author}{\bibinfo{person}{Yoshua Bengio} {and}
  \bibinfo{person}{Jean-S{\'e}bastien Sen{\'e}cal}.}
  \bibinfo{year}{2008}\natexlab{}.
\newblock \showarticletitle{Adaptive importance sampling to accelerate training
  of a neural probabilistic language model}.
\newblock \bibinfo{journal}{\emph{IEEE Transactions on Neural Networks}}
  \bibinfo{volume}{19}, \bibinfo{number}{4} (\bibinfo{year}{2008}),
  \bibinfo{pages}{713--722}.
\newblock


\bibitem[\protect\citeauthoryear{Bose, Ling, and Cao}{Bose
  et~al\mbox{.}}{2018}]%
        {bose2018adversarial}
\bibfield{author}{\bibinfo{person}{Avishek~Joey Bose}, \bibinfo{person}{Huan
  Ling}, {and} \bibinfo{person}{Yanshuai Cao}.}
  \bibinfo{year}{2018}\natexlab{}.
\newblock \showarticletitle{Adversarial Contrastive Estimation}.
\newblock  (\bibinfo{year}{2018}), \bibinfo{pages}{1021--1032}.
\newblock


\bibitem[\protect\citeauthoryear{Cai and Wang}{Cai and Wang}{2018}]%
        {cai2017kbgan}
\bibfield{author}{\bibinfo{person}{Liwei Cai} {and}
  \bibinfo{person}{William~Yang Wang}.} \bibinfo{year}{2018}\natexlab{}.
\newblock \showarticletitle{KBGAN: Adversarial Learning for Knowledge Graph
  Embeddings}. In \bibinfo{booktitle}{\emph{NAACL-HLT‘18}}.
  \bibinfo{pages}{1470--1480}.
\newblock


\bibitem[\protect\citeauthoryear{Caselles-Dupr{\'e}, Lesaint, and
  Royo-Letelier}{Caselles-Dupr{\'e} et~al\mbox{.}}{2018}]%
        {caselles2018word2vec}
\bibfield{author}{\bibinfo{person}{Hugo Caselles-Dupr{\'e}},
  \bibinfo{person}{Florian Lesaint}, {and} \bibinfo{person}{Jimena
  Royo-Letelier}.} \bibinfo{year}{2018}\natexlab{}.
\newblock \showarticletitle{Word2vec applied to recommendation: Hyperparameters
  matter}. In \bibinfo{booktitle}{\emph{RecSys'18}}. ACM,
  \bibinfo{pages}{352--356}.
\newblock


\bibitem[\protect\citeauthoryear{Chen, Ma, and Xiao}{Chen
  et~al\mbox{.}}{2018}]%
        {fastgcn}
\bibfield{author}{\bibinfo{person}{Jie Chen}, \bibinfo{person}{Tengfei Ma},
  {and} \bibinfo{person}{Cao Xiao}.} \bibinfo{year}{2018}\natexlab{}.
\newblock \showarticletitle{FastGCN: fast learning with graph convolutional
  networks via importance sampling}.
\newblock \bibinfo{journal}{\emph{ICLR'18}} (\bibinfo{year}{2018}).
\newblock


\bibitem[\protect\citeauthoryear{Chib and Greenberg}{Chib and
  Greenberg}{1995}]%
        {chib1995understanding}
\bibfield{author}{\bibinfo{person}{Siddhartha Chib} {and}
  \bibinfo{person}{Edward Greenberg}.} \bibinfo{year}{1995}\natexlab{}.
\newblock \showarticletitle{Understanding the metropolis-hastings algorithm}.
\newblock \bibinfo{journal}{\emph{The american statistician}}
  \bibinfo{volume}{49}, \bibinfo{number}{4} (\bibinfo{year}{1995}),
  \bibinfo{pages}{327--335}.
\newblock


\bibitem[\protect\citeauthoryear{Cremonesi, Koren, and Turrin}{Cremonesi
  et~al\mbox{.}}{2010}]%
        {hit1}
\bibfield{author}{\bibinfo{person}{Paolo Cremonesi}, \bibinfo{person}{Yehuda
  Koren}, {and} \bibinfo{person}{Roberto Turrin}.}
  \bibinfo{year}{2010}\natexlab{}.
\newblock \showarticletitle{Performance of recommender algorithms on top-n
  recommendation tasks}. In \bibinfo{booktitle}{\emph{Proceedings of the fourth
  ACM conference on Recommender systems}}. ACM, \bibinfo{pages}{39--46}.
\newblock


\bibitem[\protect\citeauthoryear{Ding, Tang, and Zhang}{Ding
  et~al\mbox{.}}{2018}]%
        {ding2018semi}
\bibfield{author}{\bibinfo{person}{Ming Ding}, \bibinfo{person}{Jie Tang},
  {and} \bibinfo{person}{Jie Zhang}.} \bibinfo{year}{2018}\natexlab{}.
\newblock \showarticletitle{Semi-supervised learning on graphs with generative
  adversarial nets}. In \bibinfo{booktitle}{\emph{CIKM'18}}. ACM,
  \bibinfo{pages}{913--922}.
\newblock


\bibitem[\protect\citeauthoryear{Fan, Chang, Hsieh, Wang, and Lin}{Fan
  et~al\mbox{.}}{2008}]%
        {fan2008liblinear}
\bibfield{author}{\bibinfo{person}{Rong-En Fan}, \bibinfo{person}{Kai-Wei
  Chang}, \bibinfo{person}{Cho-Jui Hsieh}, \bibinfo{person}{Xiang-Rui Wang},
  {and} \bibinfo{person}{Chih-Jen Lin}.} \bibinfo{year}{2008}\natexlab{}.
\newblock \showarticletitle{LIBLINEAR: A library for large linear
  classification}.
\newblock \bibinfo{journal}{\emph{Journal of machine learning research}}
  \bibinfo{volume}{9}, \bibinfo{number}{Aug} (\bibinfo{year}{2008}),
  \bibinfo{pages}{1871--1874}.
\newblock


\bibitem[\protect\citeauthoryear{Gao and Huang}{Gao and Huang}{2018}]%
        {gao2018self-paced}
\bibfield{author}{\bibinfo{person}{Hongchang Gao} {and} \bibinfo{person}{Heng
  Huang}.} \bibinfo{year}{2018}\natexlab{}.
\newblock \showarticletitle{Self-Paced Network Embedding}.
\newblock  (\bibinfo{year}{2018}), \bibinfo{pages}{1406--1415}.
\newblock


\bibitem[\protect\citeauthoryear{Grover and Leskovec}{Grover and
  Leskovec}{2016}]%
        {node2vec}
\bibfield{author}{\bibinfo{person}{Aditya Grover} {and} \bibinfo{person}{Jure
  Leskovec}.} \bibinfo{year}{2016}\natexlab{}.
\newblock \showarticletitle{node2vec: Scalable feature learning for networks}.
  In \bibinfo{booktitle}{\emph{KDD'16}}. ACM, \bibinfo{pages}{855--864}.
\newblock


\bibitem[\protect\citeauthoryear{Gutmann and Hyv{\"a}rinen}{Gutmann and
  Hyv{\"a}rinen}{2012}]%
        {gutmann2012noise}
\bibfield{author}{\bibinfo{person}{Michael~U Gutmann} {and}
  \bibinfo{person}{Aapo Hyv{\"a}rinen}.} \bibinfo{year}{2012}\natexlab{}.
\newblock \showarticletitle{Noise-contrastive estimation of unnormalized
  statistical models, with applications to natural image statistics}.
\newblock \bibinfo{journal}{\emph{Journal of Machine Learning Research}}
  \bibinfo{volume}{13}, \bibinfo{number}{Feb} (\bibinfo{year}{2012}),
  \bibinfo{pages}{307--361}.
\newblock


\bibitem[\protect\citeauthoryear{Hamilton, Ying, and Leskovec}{Hamilton
  et~al\mbox{.}}{2017}]%
        {graphsage}
\bibfield{author}{\bibinfo{person}{Will Hamilton}, \bibinfo{person}{Zhitao
  Ying}, {and} \bibinfo{person}{Jure Leskovec}.}
  \bibinfo{year}{2017}\natexlab{}.
\newblock \showarticletitle{Inductive representation learning on large graphs}.
  In \bibinfo{booktitle}{\emph{NIPS'17}}. \bibinfo{pages}{1024--1034}.
\newblock


\bibitem[\protect\citeauthoryear{Hsu and Lachenbruch}{Hsu and
  Lachenbruch}{2007}]%
        {hsu2007paired}
\bibfield{author}{\bibinfo{person}{Henry Hsu} {and} \bibinfo{person}{Peter~A
  Lachenbruch}.} \bibinfo{year}{2007}\natexlab{}.
\newblock \showarticletitle{Paired t test}.
\newblock \bibinfo{journal}{\emph{Wiley encyclopedia of clinical trials}}
  (\bibinfo{year}{2007}), \bibinfo{pages}{1--3}.
\newblock


\bibitem[\protect\citeauthoryear{Hu, Koren, and Volinsky}{Hu
  et~al\mbox{.}}{2008}]%
        {hu2008collaborative}
\bibfield{author}{\bibinfo{person}{Yifan Hu}, \bibinfo{person}{Yehuda Koren},
  {and} \bibinfo{person}{Chris Volinsky}.} \bibinfo{year}{2008}\natexlab{}.
\newblock \showarticletitle{Collaborative filtering for implicit feedback
  datasets}. In \bibinfo{booktitle}{\emph{ICDM'08}}. Ieee,
  \bibinfo{pages}{263--272}.
\newblock


\bibitem[\protect\citeauthoryear{Huang, Tang, Wu, Liu, and Fu}{Huang
  et~al\mbox{.}}{2014}]%
        {huang2014mining}
\bibfield{author}{\bibinfo{person}{Hong Huang}, \bibinfo{person}{Jie Tang},
  \bibinfo{person}{Sen Wu}, \bibinfo{person}{Lu Liu}, {and}
  \bibinfo{person}{Xiaoming Fu}.} \bibinfo{year}{2014}\natexlab{}.
\newblock \showarticletitle{Mining triadic closure patterns in social
  networks}. In \bibinfo{booktitle}{\emph{WWW'14}}. \bibinfo{pages}{499--504}.
\newblock


\bibitem[\protect\citeauthoryear{Kipf and Welling}{Kipf and Welling}{2017}]%
        {gcn}
\bibfield{author}{\bibinfo{person}{Thomas~N Kipf} {and} \bibinfo{person}{Max
  Welling}.} \bibinfo{year}{2017}\natexlab{}.
\newblock \showarticletitle{Semi-supervised classification with graph
  convolutional networks}.
\newblock \bibinfo{journal}{\emph{ICLR'17}} (\bibinfo{year}{2017}).
\newblock


\bibitem[\protect\citeauthoryear{Leskovec, Kleinberg, and Faloutsos}{Leskovec
  et~al\mbox{.}}{2007}]%
        {leskovec2007graph}
\bibfield{author}{\bibinfo{person}{Jure Leskovec}, \bibinfo{person}{Jon
  Kleinberg}, {and} \bibinfo{person}{Christos Faloutsos}.}
  \bibinfo{year}{2007}\natexlab{}.
\newblock \showarticletitle{Graph evolution: Densification and shrinking
  diameters}.
\newblock \bibinfo{journal}{\emph{TKDD'07}} \bibinfo{volume}{1},
  \bibinfo{number}{1} (\bibinfo{year}{2007}), \bibinfo{pages}{2--es}.
\newblock


\bibitem[\protect\citeauthoryear{Levy and Goldberg}{Levy and Goldberg}{2014}]%
        {levy2014neural}
\bibfield{author}{\bibinfo{person}{Omer Levy} {and} \bibinfo{person}{Yoav
  Goldberg}.} \bibinfo{year}{2014}\natexlab{}.
\newblock \showarticletitle{Neural word embedding as implicit matrix
  factorization}. In \bibinfo{booktitle}{\emph{NIPS'14}}.
  \bibinfo{pages}{2177--2185}.
\newblock


\bibitem[\protect\citeauthoryear{Li, Han, and Wu}{Li et~al\mbox{.}}{2018}]%
        {li2018deeper}
\bibfield{author}{\bibinfo{person}{Qimai Li}, \bibinfo{person}{Zhichao Han},
  {and} \bibinfo{person}{Xiao-Ming Wu}.} \bibinfo{year}{2018}\natexlab{}.
\newblock \showarticletitle{Deeper insights into graph convolutional networks
  for semi-supervised learning}. In \bibinfo{booktitle}{\emph{AAAI'18}}.
\newblock


\bibitem[\protect\citeauthoryear{Linden, Smith, and York}{Linden
  et~al\mbox{.}}{2003}]%
        {linden2003amazon}
\bibfield{author}{\bibinfo{person}{Greg Linden}, \bibinfo{person}{Brent Smith},
  {and} \bibinfo{person}{Jeremy York}.} \bibinfo{year}{2003}\natexlab{}.
\newblock \showarticletitle{Amazon. com recommendations: Item-to-item
  collaborative filtering}.
\newblock \bibinfo{journal}{\emph{IEEE Internet computing}}
  \bibinfo{volume}{7}, \bibinfo{number}{1} (\bibinfo{year}{2003}),
  \bibinfo{pages}{76--80}.
\newblock


\bibitem[\protect\citeauthoryear{McAuley, Targett, Shi, and Van
  Den~Hengel}{McAuley et~al\mbox{.}}{2015}]%
        {amazon}
\bibfield{author}{\bibinfo{person}{Julian McAuley},
  \bibinfo{person}{Christopher Targett}, \bibinfo{person}{Qinfeng Shi}, {and}
  \bibinfo{person}{Anton Van Den~Hengel}.} \bibinfo{year}{2015}\natexlab{}.
\newblock \showarticletitle{Image-based recommendations on styles and
  substitutes}. In \bibinfo{booktitle}{\emph{SIGIR'15}}. ACM,
  \bibinfo{pages}{43--52}.
\newblock


\bibitem[\protect\citeauthoryear{Metropolis, Rosenbluth, Rosenbluth, Teller,
  and Teller}{Metropolis et~al\mbox{.}}{1953}]%
        {metropolis1953equation}
\bibfield{author}{\bibinfo{person}{Nicholas Metropolis},
  \bibinfo{person}{Arianna~W Rosenbluth}, \bibinfo{person}{Marshall~N
  Rosenbluth}, \bibinfo{person}{Augusta~H Teller}, {and}
  \bibinfo{person}{Edward Teller}.} \bibinfo{year}{1953}\natexlab{}.
\newblock \showarticletitle{Equation of state calculations by fast computing
  machines}.
\newblock \bibinfo{journal}{\emph{The journal of chemical physics}}
  \bibinfo{volume}{21}, \bibinfo{number}{6} (\bibinfo{year}{1953}),
  \bibinfo{pages}{1087--1092}.
\newblock


\bibitem[\protect\citeauthoryear{Mikolov, Sutskever, Chen, Corrado, and
  Dean}{Mikolov et~al\mbox{.}}{2013}]%
        {word2vec}
\bibfield{author}{\bibinfo{person}{Tomas Mikolov}, \bibinfo{person}{Ilya
  Sutskever}, \bibinfo{person}{Kai Chen}, \bibinfo{person}{Greg~S Corrado},
  {and} \bibinfo{person}{Jeff Dean}.} \bibinfo{year}{2013}\natexlab{}.
\newblock \showarticletitle{Distributed representations of words and phrases
  and their compositionality}. In \bibinfo{booktitle}{\emph{NIPS'13}}.
  \bibinfo{pages}{3111--3119}.
\newblock


\bibitem[\protect\citeauthoryear{Mnih and Kavukcuoglu}{Mnih and
  Kavukcuoglu}{2013}]%
        {mnih2013learning}
\bibfield{author}{\bibinfo{person}{Andriy Mnih} {and} \bibinfo{person}{Koray
  Kavukcuoglu}.} \bibinfo{year}{2013}\natexlab{}.
\newblock \showarticletitle{Learning word embeddings efficiently with
  noise-contrastive estimation}. In \bibinfo{booktitle}{\emph{NIPS'13}}.
  \bibinfo{pages}{2265--2273}.
\newblock


\bibitem[\protect\citeauthoryear{Pan, Zhou, Cao, Liu, Lukose, Scholz, and
  Yang}{Pan et~al\mbox{.}}{2008}]%
        {pan2008one}
\bibfield{author}{\bibinfo{person}{Rong Pan}, \bibinfo{person}{Yunhong Zhou},
  \bibinfo{person}{Bin Cao}, \bibinfo{person}{Nathan~N Liu},
  \bibinfo{person}{Rajan Lukose}, \bibinfo{person}{Martin Scholz}, {and}
  \bibinfo{person}{Qiang Yang}.} \bibinfo{year}{2008}\natexlab{}.
\newblock \showarticletitle{One-class collaborative filtering}. In
  \bibinfo{booktitle}{\emph{ICDM'08}}. IEEE, \bibinfo{pages}{502--511}.
\newblock


\bibitem[\protect\citeauthoryear{Perozzi, Al-Rfou, and Skiena}{Perozzi
  et~al\mbox{.}}{2014}]%
        {deepwalk}
\bibfield{author}{\bibinfo{person}{Bryan Perozzi}, \bibinfo{person}{Rami
  Al-Rfou}, {and} \bibinfo{person}{Steven Skiena}.}
  \bibinfo{year}{2014}\natexlab{}.
\newblock \showarticletitle{Deepwalk: Online learning of social
  representations}. In \bibinfo{booktitle}{\emph{KDD'14}}. ACM,
  \bibinfo{pages}{701--710}.
\newblock


\bibitem[\protect\citeauthoryear{Qiu, Dong, Ma, Li, Wang, and Tang}{Qiu
  et~al\mbox{.}}{2018}]%
        {qiu2018network}
\bibfield{author}{\bibinfo{person}{Jiezhong Qiu}, \bibinfo{person}{Yuxiao
  Dong}, \bibinfo{person}{Hao Ma}, \bibinfo{person}{Jian Li},
  \bibinfo{person}{Kuansan Wang}, {and} \bibinfo{person}{Jie Tang}.}
  \bibinfo{year}{2018}\natexlab{}.
\newblock \showarticletitle{Network embedding as matrix factorization: Unifying
  deepwalk, line, pte, and node2vec}. In \bibinfo{booktitle}{\emph{WSDM'18}}.
  ACM, \bibinfo{pages}{459--467}.
\newblock


\bibitem[\protect\citeauthoryear{Rendle, Freudenthaler, Gantner, and
  Schmidt-Thieme}{Rendle et~al\mbox{.}}{2009}]%
        {rendle2009bpr}
\bibfield{author}{\bibinfo{person}{Steffen Rendle}, \bibinfo{person}{Christoph
  Freudenthaler}, \bibinfo{person}{Zeno Gantner}, {and} \bibinfo{person}{Lars
  Schmidt-Thieme}.} \bibinfo{year}{2009}\natexlab{}.
\newblock \showarticletitle{BPR: Bayesian personalized ranking from implicit
  feedback}. In \bibinfo{booktitle}{\emph{UAI'09}}. AUAI Press,
  \bibinfo{pages}{452--461}.
\newblock


\bibitem[\protect\citeauthoryear{Sugiyama and Kan}{Sugiyama and Kan}{2010}]%
        {mrr}
\bibfield{author}{\bibinfo{person}{Kazunari Sugiyama} {and}
  \bibinfo{person}{Min-Yen Kan}.} \bibinfo{year}{2010}\natexlab{}.
\newblock \showarticletitle{Scholarly paper recommendation via user's recent
  research interests}. In \bibinfo{booktitle}{\emph{JCDL'10}}. ACM,
  \bibinfo{pages}{29--38}.
\newblock


\bibitem[\protect\citeauthoryear{Sun, Deng, Nie, and Tang}{Sun
  et~al\mbox{.}}{2019}]%
        {sun2019rotate}
\bibfield{author}{\bibinfo{person}{Zhiqing Sun}, \bibinfo{person}{Zhi-Hong
  Deng}, \bibinfo{person}{Jian-Yun Nie}, {and} \bibinfo{person}{Jian Tang}.}
  \bibinfo{year}{2019}\natexlab{}.
\newblock \showarticletitle{Rotate: Knowledge graph embedding by relational
  rotation in complex space}.
\newblock \bibinfo{journal}{\emph{arXiv preprint arXiv:1902.10197}}
  (\bibinfo{year}{2019}).
\newblock


\bibitem[\protect\citeauthoryear{Tang, Qu, Wang, Zhang, Yan, and Mei}{Tang
  et~al\mbox{.}}{2015}]%
        {line}
\bibfield{author}{\bibinfo{person}{Jian Tang}, \bibinfo{person}{Meng Qu},
  \bibinfo{person}{Mingzhe Wang}, \bibinfo{person}{Ming Zhang},
  \bibinfo{person}{Jun Yan}, {and} \bibinfo{person}{Qiaozhu Mei}.}
  \bibinfo{year}{2015}\natexlab{}.
\newblock \showarticletitle{Line: Large-scale information network embedding}.
  In \bibinfo{booktitle}{\emph{WWW'15}}. \bibinfo{pages}{1067--1077}.
\newblock


\bibitem[\protect\citeauthoryear{Tu, Liu, Liu, and Sun}{Tu
  et~al\mbox{.}}{2017}]%
        {tu2017cane}
\bibfield{author}{\bibinfo{person}{Cunchao Tu}, \bibinfo{person}{Han Liu},
  \bibinfo{person}{Zhiyuan Liu}, {and} \bibinfo{person}{Maosong Sun}.}
  \bibinfo{year}{2017}\natexlab{}.
\newblock \showarticletitle{Cane: Context-aware network embedding for relation
  modeling}. In \bibinfo{booktitle}{\emph{ACL'17}}.
  \bibinfo{pages}{1722--1731}.
\newblock


\bibitem[\protect\citeauthoryear{Veli{\v{c}}kovi{\'c}, Cucurull, Casanova,
  Romero, Lio, and Bengio}{Veli{\v{c}}kovi{\'c} et~al\mbox{.}}{2018}]%
        {velivckovic2017graph}
\bibfield{author}{\bibinfo{person}{Petar Veli{\v{c}}kovi{\'c}},
  \bibinfo{person}{Guillem Cucurull}, \bibinfo{person}{Arantxa Casanova},
  \bibinfo{person}{Adriana Romero}, \bibinfo{person}{Pietro Lio}, {and}
  \bibinfo{person}{Yoshua Bengio}.} \bibinfo{year}{2018}\natexlab{}.
\newblock \showarticletitle{Graph attention networks}.
\newblock \bibinfo{journal}{\emph{ICLR'18}} (\bibinfo{year}{2018}).
\newblock


\bibitem[\protect\citeauthoryear{Wang, Yu, Zhang, Gong, Xu, Wang, Zhang, and
  Zhang}{Wang et~al\mbox{.}}{2017b}]%
        {wang2017irgan}
\bibfield{author}{\bibinfo{person}{Jun Wang}, \bibinfo{person}{Lantao Yu},
  \bibinfo{person}{Weinan Zhang}, \bibinfo{person}{Yu Gong},
  \bibinfo{person}{Yinghui Xu}, \bibinfo{person}{Benyou Wang},
  \bibinfo{person}{Peng Zhang}, {and} \bibinfo{person}{Dell Zhang}.}
  \bibinfo{year}{2017}\natexlab{b}.
\newblock \showarticletitle{Irgan: A minimax game for unifying generative and
  discriminative information retrieval models}. In
  \bibinfo{booktitle}{\emph{SIGIR'17}}. ACM, \bibinfo{pages}{515--524}.
\newblock


\bibitem[\protect\citeauthoryear{Wang, Yin, Hu, Lian, Wang, and Huang}{Wang
  et~al\mbox{.}}{2018}]%
        {wang2018neural}
\bibfield{author}{\bibinfo{person}{Qinyong Wang}, \bibinfo{person}{Hongzhi
  Yin}, \bibinfo{person}{Zhiting Hu}, \bibinfo{person}{Defu Lian},
  \bibinfo{person}{Hao Wang}, {and} \bibinfo{person}{Zi Huang}.}
  \bibinfo{year}{2018}\natexlab{}.
\newblock \showarticletitle{Neural memory streaming recommender networks with
  adversarial training}. In \bibinfo{booktitle}{\emph{KDD'18}}. ACM,
  \bibinfo{pages}{2467--2475}.
\newblock


\bibitem[\protect\citeauthoryear{Wang, Cui, Wang, Pei, Zhu, and Yang}{Wang
  et~al\mbox{.}}{2017a}]%
        {wang2017community}
\bibfield{author}{\bibinfo{person}{Xiao Wang}, \bibinfo{person}{Peng Cui},
  \bibinfo{person}{Jing Wang}, \bibinfo{person}{Jian Pei},
  \bibinfo{person}{Wenwu Zhu}, {and} \bibinfo{person}{Shiqiang Yang}.}
  \bibinfo{year}{2017}\natexlab{a}.
\newblock \showarticletitle{Community preserving network embedding}. In
  \bibinfo{booktitle}{\emph{AAAI'17}}.
\newblock


\bibitem[\protect\citeauthoryear{Weston, Bengio, and Usunier}{Weston
  et~al\mbox{.}}{2011}]%
        {weston2011wsabie}
\bibfield{author}{\bibinfo{person}{Jason Weston}, \bibinfo{person}{Samy
  Bengio}, {and} \bibinfo{person}{Nicolas Usunier}.}
  \bibinfo{year}{2011}\natexlab{}.
\newblock \showarticletitle{Wsabie: Scaling up to large vocabulary image
  annotation}. In \bibinfo{booktitle}{\emph{IJCAI'11}}.
\newblock


\bibitem[\protect\citeauthoryear{Xu, Hu, Leskovec, and Jegelka}{Xu
  et~al\mbox{.}}{2018}]%
        {xu2018powerful}
\bibfield{author}{\bibinfo{person}{Keyulu Xu}, \bibinfo{person}{Weihua Hu},
  \bibinfo{person}{Jure Leskovec}, {and} \bibinfo{person}{Stefanie Jegelka}.}
  \bibinfo{year}{2018}\natexlab{}.
\newblock \showarticletitle{How powerful are graph neural networks?}
\newblock \bibinfo{journal}{\emph{arXiv preprint arXiv:1810.00826}}
  (\bibinfo{year}{2018}).
\newblock


\bibitem[\protect\citeauthoryear{Ying, He, Chen, Eksombatchai, Hamilton, and
  Leskovec}{Ying et~al\mbox{.}}{2018}]%
        {pinsage}
\bibfield{author}{\bibinfo{person}{Rex Ying}, \bibinfo{person}{Ruining He},
  \bibinfo{person}{Kaifeng Chen}, \bibinfo{person}{Pong Eksombatchai},
  \bibinfo{person}{William~L Hamilton}, {and} \bibinfo{person}{Jure Leskovec}.}
  \bibinfo{year}{2018}\natexlab{}.
\newblock \showarticletitle{Graph convolutional neural networks for web-scale
  recommender systems}. In \bibinfo{booktitle}{\emph{KDD'18}}. ACM,
  \bibinfo{pages}{974--983}.
\newblock


\bibitem[\protect\citeauthoryear{Zhang, Chen, Wang, and Yu}{Zhang
  et~al\mbox{.}}{2013}]%
        {dns}
\bibfield{author}{\bibinfo{person}{Weinan Zhang}, \bibinfo{person}{Tianqi
  Chen}, \bibinfo{person}{Jun Wang}, {and} \bibinfo{person}{Yong Yu}.}
  \bibinfo{year}{2013}\natexlab{}.
\newblock \showarticletitle{Optimizing Top-N Collaborative Filtering via
  Dynamic Negative Item Sampling}. In \bibinfo{booktitle}{\emph{SIGIR'13}}.
  ACM, \bibinfo{pages}{785--788}.
\newblock


\bibitem[\protect\citeauthoryear{Zhang, Yao, Shao, and Chen}{Zhang
  et~al\mbox{.}}{2019}]%
        {zhang2019nscaching}
\bibfield{author}{\bibinfo{person}{Yongqi Zhang}, \bibinfo{person}{Quanming
  Yao}, \bibinfo{person}{Yingxia Shao}, {and} \bibinfo{person}{Lei Chen}.}
  \bibinfo{year}{2019}\natexlab{}.
\newblock \showarticletitle{NSCaching: Simple and Efficient Negative Sampling
  for Knowledge Graph Embedding}.
\newblock  (\bibinfo{year}{2019}), \bibinfo{pages}{614--625}.
\newblock


\bibitem[\protect\citeauthoryear{Zhang and Zweigenbaum}{Zhang and
  Zweigenbaum}{2018}]%
        {zhang2018gneg}
\bibfield{author}{\bibinfo{person}{Zheng Zhang} {and} \bibinfo{person}{Pierre
  Zweigenbaum}.} \bibinfo{year}{2018}\natexlab{}.
\newblock \showarticletitle{GNEG: Graph-Based Negative Sampling for word2vec}.
  In \bibinfo{booktitle}{\emph{ACL'18}}. \bibinfo{pages}{566--571}.
\newblock


\bibitem[\protect\citeauthoryear{Zhao, McAuley, and King}{Zhao
  et~al\mbox{.}}{2015}]%
        {zhao2015improving}
\bibfield{author}{\bibinfo{person}{Tong Zhao}, \bibinfo{person}{Julian
  McAuley}, {and} \bibinfo{person}{Irwin King}.}
  \bibinfo{year}{2015}\natexlab{}.
\newblock \showarticletitle{Improving latent factor models via personalized
  feature projection for one class recommendation}. In
  \bibinfo{booktitle}{\emph{CIKM'15}}. ACM, \bibinfo{pages}{821--830}.
\newblock


\bibitem[\protect\citeauthoryear{Zhou, Liu, Liu, Liu, and Gao}{Zhou
  et~al\mbox{.}}{2017}]%
        {zhou2017scalable}
\bibfield{author}{\bibinfo{person}{Chang Zhou}, \bibinfo{person}{Yuqiong Liu},
  \bibinfo{person}{Xiaofei Liu}, \bibinfo{person}{Zhongyi Liu}, {and}
  \bibinfo{person}{Jun Gao}.} \bibinfo{year}{2017}\natexlab{}.
\newblock \showarticletitle{Scalable graph embedding for asymmetric proximity}.
  In \bibinfo{booktitle}{\emph{AAAI'17}}.
\newblock


\bibitem[\protect\citeauthoryear{Zhou, Ma, Zhang, Zhou, and Yang}{Zhou
  et~al\mbox{.}}{2020}]%
        {zhou2020contrastive}
\bibfield{author}{\bibinfo{person}{Chang Zhou}, \bibinfo{person}{Jianxin Ma},
  \bibinfo{person}{Jianwei Zhang}, \bibinfo{person}{Jingren Zhou}, {and}
  \bibinfo{person}{Hongxia Yang}.} \bibinfo{year}{2020}\natexlab{}.
\newblock \bibinfo{title}{Contrastive Learning for Debiased Candidate
  Generation in Large-Scale Recommender Systems}.
\newblock
\newblock
\showeprint[arxiv]{cs.IR/2005.12964}


\end{thebibliography}
\appendix
\clearpage
\section{Appendix}
In the appendix, we first report the implementation notes of \model, including the running environment and implementation details. Then, the encoder algorithms are presented. Next, the detailed description of each baseline is provided. Finally, we present datasets, evaluation metrics and the DFS algorithm in detail.
\subsection{Implementation Notes}

\noindent \textbf{Running Environment.} The experiments are conducted on a single Linux server with 14 Intel(R) Xeon(R) CPU E5-268 v4 @ 2.40GHz, 256G RAM and 8 NVIDIA GeForce RTX 2080TI-11GB. Our proposed \model \  is implemented in Tensorflow 1.14 and Python 3.7. 

\noindent \textbf{Implementation Details.} All algorithms can be divided into three parts: \textit{negative sampling}, \textit{positive sampling} and \textit{embedding learning}. In this paper, we focus on negative sampling strategy. A negative sampler selects a negative sample for each positive pair and feeds it with a corresponding positive pair to the encoder to learn node embeddings. With the exception of PinSAGE, the ratio of positive to negative pairs is set to $1:1$ for all negative sampling strategies. PinSAGE uniformly selects 20 negative samples per batch and adds 5 ``hard negative samples'' per positive pair. Nodes ranked at top-100 according to PageRank scores are randomly sampled as ``hard negative samples''. The candidate size $S$ in DNS is set to 5 for all datasets. The value of $T$ in KBGAN is set to 200 for Amazon and Alibaba datasets, and 100 for other datasets.

For recommendation task, we evenly divide the edges in each dataset into 11 parts, 10 for ten-fold cross validation and the extra one as a validation set to determine hyperparameters.
For recommendation datasets, the node types contain $U$ and $I$ representing \textit{user} and \textit{item} respectively. Thus, the paths for information aggregation are set to $U-I-U$ and $I-U-I$ for GraphSAGE and GCN encoders.
All the important parameters in \smodel are determined by a rough grid search. For example, the learning rate is the best among [$10^{-5}$,$10^{-4}$,$10^{-3}$,$10^{-2}$]. Adam with $\beta_1=0.9, \beta_2=0.999$ serves as the optimizer for all the experiments. The values of hyperparameters in different datasets are listed as follows:

\begin{center}
\begin{tabular}{c|c|c|c}
\hline 
     &MovieLens & Amazon & Alibaba  \\
     \hline
learning rate & $10^{-3}$ & $10^{-3}$ & $10^{-3}$\\
embedding dim& 256 & 512 &512\\
margin $\gamma$ & 0.1 & 0.1 & 0.1\\
batch size& 256 & 512 & 512\\
$M$ in Hits@k& ALL & 500 & 500\\
\hline
\end{tabular}
\end{center}

For link prediction task, we employ five-fold cross validation to train all methods. The embedding dimension is set to 256, and the margin $\gamma$ also be set to 0.1 for all strategies. The model parameters are updated and optimized by stochastic gradient descent with Adam optimizer.  

In multi-label classification task, we first employ all the edges to learn node embeddings. Then, these embeddings serve as features  to train the classifier. Embedding dimension is set to 128 and the margin $\gamma$ is set as 0.1 on the BlogCatalog dataset for all negative sampling strategies. The optimizer also adopts Adam updating rule.

\subsection{Encoders}\label{app:encoders} 
We perform experiments on three algorithms as follows. 
\begin{itemize}
    \item \textbf{DeepWalk}~\cite{deepwalk} is the most representative graph representation algorithm. According to the Skip-gram model~\cite{word2vec}, DeepWalk samples nodes co-occurring in the same window in random-walk paths as positive pairs, which acts as a data augmentation. This strategy serves as ``sampling from $\hat{p_d}$'' in the view of SampledNCE (\S\ref{unify}). 
    

    \item \textbf{GCN}~\cite{gcn} introduces a simple and effective neighbor aggregation operation for graph representation learning. Although GCN samples positive pairs from direct edges in the graph, the final embeddings of nodes are implicitly constrained by the smoothing effect of graph convolution. However, due to its poor scalability and high memory consumption, GCN cannot handle large graphs.
    \item \textbf{GraphSAGE}~\cite{graphsage} is an inductive framework of graph convolution. A fixed number of neighbors are sampled to make the model suitable for batch training, bringing about its prevalence in large-scale graphs. The \textbf{mean}-aggregator is used in our experiments.
\end{itemize}

\subsection{Baselines}\label{app:baselines}
We give detailed negative sampling strategies as follows. For each encoder, the hyperparameters for various negative sampling strategies remain the same. 
\begin{itemize}
    \item\textbf{Power of Degree}~\cite{word2vec}. This strategy is first proposed in word2vec, in which $p_n(v)\propto deg(v)^\beta$. In word embedding, the best $\alpha$ is found to be 0.75 by experiments and followed by most graph embedding algorithms~\cite{line,deepwalk}. We use $Deg^{0.75}$ to denote this widely-used setting.
    
    \item\textbf{RNS}~\cite{caselles2018word2vec}. RNS means sampling negative nodes at uniform random. Previous work~\cite{caselles2018word2vec} mentions that in recommendation, RNS performs better and more robust than $Deg^{0.75}$ during tuning hyperparameters. This phenomenon is also verified in our recommendation experiments by using this setting as a baseline. 
    
    \item\textbf{WRMF}~\cite{hu2008collaborative,pan2008one}. Weighted Regularized Matrix Factorization is an implicit MF model, which adapts a weight to reduce the impact of  unobserved interactions and utilizes an altering-least-squares optimization process. To solve the high computational costs, WRMF proposes three negative sampling schemes, including Uniform Random Sampling (RNS), Item-Oriented Sampling (ItemPop), User-Oriented Sampling (Unconsider in our paper). 

    \item\textbf{DNS}~\cite{dns}. Dynamic Negative Sampling (DNS) is originally designed for sampling negative items to improve pair-wise collaborative filtering. For each user $u$, DNS samples negative $S$ items $\{v_0, ..., v_{S-1}\}$ and only retains the one with the highest scoring function $s(u, v)$. DNS can be applied to graph-based recommendation by replacing $s(u,v)$ in BPR~\cite{rendle2009bpr} with the inner product of embedding $\!u\!v^\top$. 

    \item\textbf{PinSAGE}~\cite{pinsage}. In PinSAGE, a technique called ``hard negative items'' is introduced to improve its performance. The hard negative items refer to the high ranking items according to users' Personalized PageRank scores. Finally, negative items are sampled from top-K hard negative items.
    
    \item\textbf{WARP}~\cite{weston2011wsabie,zhao2015improving}. Weighted Approximate-Rank Pairwise(WARP) utilizes SGD and a novel sampling trick to approximate ranks. For each positive pair $(u,v)$, WARP proposes uniform sampling with rejection strategy to find a negative item $v'$ until $\!u\!v'^\top - \!u\!v^\top + \gamma > 0$.

    \item\textbf{IRGAN}~\cite{wang2017irgan}. IRGAN is a GAN-based IR model that unify  generative and discriminative information retrieval models. The generator generates adversarial negative samples to fool the discriminator while the discriminator tries to distinguish them from true positive samples.  
    
    \item\textbf{KBGAN}~\cite{cai2017kbgan}. KBGAN proposed a adversarial sampler to improve the performances of knowledge graph embedding models. To reduce computational complexity, KBGAN uniformly randomly selects $T$ nodes to calculate the probability distribution for generating negative samples. Inspired by it, NMRN \cite{wang2018neural} employed a GAN-based adversarial training framework in streaming recommender model. 
\end{itemize}

\subsection{Datasets}\label{app:dataset}
The detailed descriptions for five datasets are listed as follows.
\begin{itemize}
    \item \textbf{MovieLens\footnote{https://grouplens.org/datasets/movielens/100k/}} is a widely-used movie rating dataset for evaluating recommender systems. We choose the MovieLens-100k version that contains 100,000 interaction records generated by 943 users on 1682 movies. 

    \item \textbf{Amazon\footnote{http://jmcauley.ucsd.edu/data/amazon/links.html}} is a large E-commerce dataset introduced in \cite{amazon}, which contains purchase records and review texts whose time stamps span from May 1996 to July 2014. 
    In the experiments, we take the data from the commonly-used ``electronics'' category to establish a user-item graph.

    \item \textbf{Alibaba\footnote{we use Alibaba dataset to represent the real data we collected.}} is constructed based on the data from another large E-commerce platform, which contains user's purchase records and items' attribute information. The data are organized as a user-item graph for recommendation.
    
    \item \textbf{Arxiv GR-QC} \cite{leskovec2007graph} is a collaboration network generated from the e-print arXiv where nodes represent authors and edges indicate collaborations between authors. 

    \item \textbf{BlogCatalog} is a network of social relationships provided by blogger authors. The labels represent the topic categories provided by the authors and each blogger may be associated to multiple categories. There are overall 39 different categories for BlogCatalog.

\end{itemize}


\subsection{Evaluation Metrics}\label{app:metrics}

\noindent \textbf{Recommendation.}
To evaluate the performance of \model \ for recommendation task,  $Hits@k$~\cite{hit1} and mean reciprocal ranking ($MRR$) ~\cite{mrr} serve as evaluation methodologies.
\begin{itemize}
    \item \textbf{Hits@k} is introduced by the classic work~\cite{mrr} to measure the performance of personalized recommendation at a coarse granularity. Specifically, we can compute $Hits@k$ of a recommendation system as follows:
\begin{enumerate}
    \item For each user-item pair $(u, v)$ in the test set $D_{test}$, we randomly select $M$ additional items  $\{v_0', ..., v_{M-1}'\}$ which are never showed to the queried user $u$. 
    \item Form a ranked list of $\{\!u\!v^\top, \!u\!v_0'^\top, ..., \!u\!v_{M-1}'^\top\}$ in descending order.
    \item Examine whether the rank of $\!u\!v^\top$ is less than $k$ (hit) or not (miss).
    \item Calculate $Hits@k$ according to $Hits@k = \frac{\# hit@k}{|D_{test}|}$.
\end{enumerate}
 
\item \textbf{MRR} is another evaluation metric for recommendation task, which measures the performance at a finer granularity than $Hits@k$. $MRR$ assigns different scores to different ranks of item $v$ in the ranked list mentioned above when querying user $u$, which is defined as follows:

\begin{equation*}
    MRR = \frac{1}{|D_{test}|}\sum_{(u_i, v_i) \in D_{test}}{\frac{1}{rank(v_i)}}.
\end{equation*}
\end{itemize}

In our experiment, we set $k$ as $\{10, 30\}$, and $M$ is set to 500 for the Alibaba and Amazon datasets. For the MovieLens dataset,  we use all unconnected items for each user as $M$ items to evaluate the hit rate.

\noindent \textbf{Link Prediction.} The standard evaluation metric AUC is widely applied to evaluate the performance on link prediction task \cite{node2vec,tu2017cane, zhou2017scalable}. Here, we utilize \textit{roc$\_$auc$\_$score} function from scikit-learn to calculate prediction score. The specific calculation process is demonstrated as follows.

\begin{enumerate}
    \item We extract $30\%$ of true edges and remove them from the whole graph while ensuring the residual graph is connected.
    \item We sample 30 $\%$ false edges, and then combine $30\%$ of true edges and false edges into a test set.
    \item Each graph embedding algorithm with various negative sampling strategies is trained using the residual graph, and then the embedding for each node is obtained.
    \item We predict the probability of a node pair being a true edge according to the inner product. We finally calculate AUC score according to the probability via  \textit{roc$\_$auc$\_$score} function.
    
\end{enumerate}

\noindent \textbf{Multi-label Classification.} In the multi-label classification, we adopt Micro-F1 and Macro-F1 as evaluation metrics, which are widely used in many previous works \cite{word2vec, node2vec}. 
In our experiments, the F1 score is calculated by scikit-learn.
In detail, we randomly select a percentage of labeled nodes, $T_R \in \{10$\%$,50$\%$,90$\%$\}$ to learn node embeddings. Then the learned embeddings  is used as features to train the classifier and predict the labels of the remaining nodes. For fairness in comparison, we used the same training set to train the classifier for all negative sampling strategies. For each training ratio, we repeat the process 10 times and report the average scores.

\subsection{DFS}
Here we introduce a DFS strategy to generate traversal sequence $L$ where adjacent nodes in the sequence is also adjacent in the graph.\label{app:dfs}
\begin{algorithm}[hb]
    \caption{DFS}
    \KwIn{
    Current node $x$.\\
    Global variables: graph $G=(V, E)$, sequence $L$.
    }
        Append $x$ to $L$\\
        \For{each unvisited neighbor $y$ of $x$ in $G$}{
            DFS($y$)\\
            Append $x$ to $L$
        }
    \end{algorithm}

\end{document}